\documentclass[10pt,journal,compsoc]{IEEEtran}
\ifCLASSOPTIONcompsoc
  \usepackage[nocompress]{cite}
\else
  \usepackage{cite}
\fi
\ifCLASSINFOpdf
\else
\fi

\usepackage{times}
\usepackage{epsfig}
\usepackage{graphicx}
\usepackage{amsmath}
\usepackage{amssymb}
\usepackage{multirow}
\usepackage{subfig}
\usepackage{amsthm}
\usepackage{bbm}
\usepackage[ruled,vlined]{algorithm2e}
\usepackage{makecell}
\usepackage{rotating}
\usepackage{booktabs}
\usepackage{float}
\usepackage{color}
\usepackage{xcolor}

\newcommand{\vect}[1]{{#1}}
\newcommand{\matr}[1]{{#1}}

\newcommand{\set}[1]{\mathcal{#1}}

\begin{document}
%
% paper title
% Titles are generally capitalized except for words such as a, an, and, as,
% at, but, by, for, in, nor, of, on, or, the, to and up, which are usually
% not capitalized unless they are the first or last word of the title.
% Linebreaks \\ can be used within to get better formatting as desired.
% Do not put math or special symbols in the title.
\title{Revisiting Realistic Test-Time Training: Sequential Inference and Adaptation by Anchored Clustering Regularized Self-Training}
%
%
% author names and IEEE memberships
% note positions of commas and nonbreaking spaces ( ~ ) LaTeX will not break
% a structure at a ~ so this keeps an author's name from being broken across
% two lines.
% use \thanks{} to gain access to the first footnote area
% a separate \thanks must be used for each paragraph as LaTeX2e's \thanks
% was not built to handle multiple paragraphs
%
%
%\IEEEcompsocitemizethanks is a special \thanks that produces the bulleted
% lists the Computer Society journals use for "first footnote" author
% affiliations. Use \IEEEcompsocthanksitem which works much like \item
% for each affiliation group. When not in compsoc mode,
% \IEEEcompsocitemizethanks becomes like \thanks and
% \IEEEcompsocthanksitem becomes a line break with idention. This
% facilitates dual compilation, although admittedly the differences in the
% desired content of \author between the different types of papers makes a
% one-size-fits-all approach a daunting prospect. For instance, compsoc 
% journal papers have the author affiliations above the "Manuscript
% received ..."  text while in non-compsoc journals this is reversed. Sigh.

\author{
        Yongyi~Su,
        Xun~Xu,
        Tianrui Li,
        Kui~Jia
        % Yongyi~Su,
        % Xun~Xu,
        % Tianrui~Li,
        % Kui~Jia
        % <-this % stops a space
\IEEEcompsocitemizethanks{\IEEEcompsocthanksitem X. Xu is with I2R, A-STAR.
Y. Su, and K. Jia are with the School of Electronic and Information Engineering, South China University of Technology. Tianrui Li is with Southwest Jiaotong University.\protect}

% note need leading \protect in front of \\ to get a newline within \thanks as
% \\ is fragile and will error, could use \hfil\break instead.

\IEEEcompsocitemizethanks{\IEEEcompsocthanksitem Y. Su and X. Xu contributed equally to this work. Corresponding to X. Xu: {alex.xun.xu}@gmail.com\protect\\
}
\thanks{}
}

% note the % following the last \IEEEmembership and also \thanks - 
% these prevent an unwanted space from occurring between the last author name
% and the end of the author line. i.e., if you had this:
% 
% \author{....lastname \thanks{...} \thanks{...} }
%                     ^------------^------------^----Do not want these spaces!
%
% a space would be appended to the last name and could cause every name on that
% line to be shifted left slightly. This is one of those "LaTeX things". For
% instance, "\textbf{A} \textbf{B}" will typeset as "A B" not "AB". To get
% "AB" then you have to do: "\textbf{A}\textbf{B}"
% \thanks is no different in this regard, so shield the last } of each \thanks
% that ends a line with a % and do not let a space in before the next \thanks.
% Spaces after \IEEEmembership other than the last one are OK (and needed) as
% you are supposed to have spaces between the names. For what it is worth,
% this is a minor point as most people would not even notice if the said evil
% space somehow managed to creep in.

% The paper headers
\markboth{Journal of \LaTeX\ Class Files,~Vol.~14, No.~8, August~2015}%
{Shell \MakeLowercase{\textit{et al.}}: Bare Demo of IEEEtran.cls for Computer Society Journals}
% The only time the second header will appear is for the odd numbered pages
% after the title page when using the twoside option.
% 

% *** Note that you probably will NOT want to include the author's ***
% *** name in the headers of peer review papers.                   ***
% You can use \ifCLASSOPTIONpeerreview for conditional compilation here if
% you desire.

% The publisher's ID mark at the bottom of the page is less important with
% Computer Society journal papers as those publications place the marks
% outside of the main text columns and, therefore, unlike regular IEEE
% journals, the available text space is not reduced by their presence.
% If you want to put a publisher's ID mark on the page you can do it like
% this:
%\IEEEpubid{0000--0000/00\$00.00~\copyright~2015 IEEE}
% or like this to get the Computer Society new two part style.
%\IEEEpubid{\makebox[\columnwidth]{\hfill 0000--0000/00/\$00.00~\copyright~2015 IEEE}%
%\hspace{\columnsep}\makebox[\columnwidth]{Published by the IEEE Computer Society\hfill}}
% Remember, if you use this you must call \IEEEpubidadjcol in the second
% column for its text to clear the IEEEpubid mark (Computer Society jorunal
% papers don't need this extra clearance.)

% use for special paper notices
%\IEEEspecialpapernotice{(Invited Paper)}

% for Computer Society papers, we must declare the abstract and index terms
% PRIOR to the title within the \IEEEtitleabstractindextext IEEEtran
% command as these need to go into the title area created by \maketitle.
% As a general rule, do not put math, special symbols or citations
% in the abstract or keywords.
\IEEEtitleabstractindextext{%
\begin{abstract}
  Deploying models on target domain data subject to distribution shift requires adaptation. Test-time training~(TTT) emerges as a solution to this adaptation under a realistic scenario where access to full source domain data is not available, and instant inference on the target domain is required. Despite many efforts into TTT, there is a confusion over the experimental settings, thus leading to unfair comparisons. In this work, we first revisit TTT assumptions and categorize TTT protocols by two key factors. Among the multiple protocols, we adopt a realistic sequential test-time training~(sTTT) protocol, under which we develop a \textit{test-time anchored clustering~(TTAC)} approach to enable stronger test-time feature learning. TTAC discovers clusters in both source and target domains and matches the target clusters to the source ones to improve adaptation. When source domain information is strictly absent~(i.e. source-free) we further develop an efficient method to infer source domain distributions for anchored clustering.
  Finally, self-training~(ST) has demonstrated great success in learning from unlabeled data and we empirically figure out that applying ST alone to TTT is prone to confirmation bias. Therefore, a more effective TTT approach is introduced by regularizing self-training with anchored clustering, and the improved model is referred to as TTAC++.
  %Pseudo label filtering and iterative updating are developed to improve the effectiveness and efficiency of anchored clustering.
  We demonstrate that, under all TTT protocols, TTAC++ consistently outperforms the state-of-the-art methods on five TTT datasets, including corrupted target domain, selected hard samples, synthetic-to-real adaptation and adversarially attacked target domain. We hope this work will provide a fair benchmarking of TTT methods, and future research should be compared within respective protocols.
\end{abstract}

% Note that keywords are not normally used for peerreview papers.
}

% make the title area
\maketitle

% To allow for easy dual compilation without having to reenter the
% abstract/keywords data, the \IEEEtitleabstractindextext text will
% not be used in maketitle, but will appear (i.e., to be "transported")
% here as \IEEEdisplaynontitleabstractindextext when the compsoc 
% or transmag modes are not selected <OR> if conference mode is selected 
% - because all conference papers position the abstract like regular
% papers do.
\IEEEdisplaynontitleabstractindextext
% \IEEEdisplaynontitleabstractindextext has no effect when using
% compsoc or transmag under a non-conference mode.

% For peer review papers, you can put extra information on the cover
% page as needed:
% \ifCLASSOPTIONpeerreview
% \begin{center} \bfseries EDICS Category: 3-BBND \end{center}
% \fi
%
% For peerreview papers, this IEEEtran command inserts a page break and
% creates the second title. It will be ignored for other modes.
\IEEEpeerreviewmaketitle

\IEEEraisesectionheading{\section{Introduction}\label{sec:introduction}}
% Computer Society journal (but not conference!) papers do something unusual
% with the very first section heading (almost always called "Introduction").
% They place it ABOVE the main text! IEEEtran.cls does not automatically do
% this for you, but you can achieve this effect with the provided
% \IEEEraisesectionheading{} command. Note the need to keep any \label that
% is to refer to the section immediately after \section in the above as
% \IEEEraisesectionheading puts \section within a raised box.

% The very first letter is a 2 line initial drop letter followed
% by the rest of the first word in caps (small caps for compsoc).
% 
% form to use if the first word consists of a single letter:
% \IEEEPARstart{A}{demo} file is ....
% 
% form to use if you need the single drop letter followed by
% normal text (unknown if ever used by the IEEE):
% \IEEEPARstart{A}{}demo file is ....
% 
% Some journals put the first two words in caps:
% \IEEEPARstart{T}{his demo} file is ....
% 
% Here we have the typical use of a "T" for an initial drop letter
% and "HIS" in caps to complete the first word.
%\IEEEPARstart{A}{ction} recognition has long been a central \section{Introduction}
 
\IEEEPARstart{T}{he} recent success in deep learning is attributed to the availability of large labeled data~\cite{krizhevsky2012imagenet,carreira2017quo,zhou2018brief} and the assumption of i.i.d. between training and test datasets. Such assumptions could be violated when testing data features a drastic difference from the training data, e.g. training on synthetic images and test on real ones, or training on clean samples and test on corrupted ones. This situation is often referred to as domain shift~\cite{quinonero2008dataset,ben2010theory,pan2009survey}. To tackle this issue, domain adaptation~(DA)~\cite{wang2018deep} emerges and the labeled training data and unlabeled testing data are often referred to as source and target data/domains respectively.
The existing DA works either require the access to both source and target domain data during training~\cite{ganin2015unsupervised,tzeng2017adversarial,hoffman2018cycada} or training on multiple domains simultaneously~\cite{zhou2021domain}. The former approaches render the methods restrictive to limited scenarios where source domain data is always available during adaptation while the latter ones are computationally more expensive.

To alleviate the dependence on source domain data, which may be inaccessible due to privacy issues or storage overhead, source-free domain adaptation (SFDA) emerges which handles DA on target data without access to source data~\cite{pmlr-v119-liang20a,kundu2020universal,yang2021generalized,xia2021adaptive,liu2021ttt++}. SFDA is often achieved through self-training~\cite{pmlr-v119-liang20a,qiu2021source}, self-supervised learning~\cite{liu2021ttt++,huang2021model} or introducing prior knowledge~\cite{pmlr-v119-liang20a} and it requires multiple training epochs on the full target {dataset} to allow model convergence. Despite easing the dependence on source data, SFDA has major drawbacks in a more realistic domain adaptation scenario where test data arrives in a stream and inference or prediction must be taken instantly, and this setting is often referred to as test-time training (TTT) or adaptation (TTA)~\cite{sun2020test,wang2020tent,iwasawa2021test,liu2021ttt++,chen2022contrastive,gandelsman2022test}. Despite the attractive feature of adaption at {test time}, we notice a confusion of what defines a test-time training and as a result comparing apples and oranges happens frequently in the community. In this work, we first categorize TTT by \textbf{two key factors} after summarizing various definitions made in existing works. First, under a realistic TTT setting, test samples are sequentially streamed and predictions should be made instantly upon the arrival of a new test sample. More specifically, the prediction of test sample $x_T$, arriving at time stamp $T$, should not be affected by any subsequent samples, $\{x_t\}_{t=T+1\cdots\infty}$. The sequential protocol widely exists in many real-world application. For example, in video surveillance, cameras are expected to function instantly after installment and adaptation to target domain must be carried out on-the-fly. Throughout this work, We refer to the sequential streaming setting as the \textbf{one-pass adaptation} protocol and any other protocols violating this assumption are referred to as \textbf{multi-pass adaptation} (model may be updated on all test data for multiple epochs before inference). Second, we notice some recent works must \textbf{modify source domain training loss}, e.g. by introducing additional self-supervised branch, to allow more effective TTT~\cite{sun2020test,liu2021ttt++}. This will introduce additional overhead in the deployment of TTT because re-training on some source dataset, e.g. ImageNet, is computationally expensive. Thus, we distinguish methods by whether source domain training objective is modified or not.
% Given the distinctions above, we provide a categorization of existing works, which are often confused under the umbrella of TTT, TTA and SFDA, and comparisons within each category are then fair.

In this work, we aim to tackle on the most realistic TTT protocol, i.e. one-pass test time training with no modifications to training objective. We refer to this new TTT protocol as \textbf{sequential test time training (sTTT)}. 
The proposed setting is similar to TTA proposed in~\cite{wang2020tent} except for not restricting access to a light-weight distribution information from the source domain. We believe having access to distribution information, e.g. distributions' mean and covariance, in the source domain is a realistic assumption for two reasons. First, the objective of TTT is efficient adaptation at {test time}, this assumption only requires storing the means and covariance matrices which are memory efficient. Moreover, feature distribution information will not pose any threat to privacy leakage as inverting backbone network, e.g. CNN, is known to be very challenging~\cite{gilbert2017towards}. Nevertheless, we are aware that under certain scenarios source domain distribution information is not always available for test-time training, i.e. \textbf{source-free test-time training}. Such a situation could happen when source distribution is not recorded during training, or model is trained through federated learning where access to the whole source data is prohibited~\cite{li2021survey}. A robust TTT method should therefore be versatile and still function well in the absence of source domain distribution information. %in the absence of source domain distribution information, we propose a technique to estimate the distribution parameters from model weights alone. %Empirical evaluation demonstrates the efficacy of learning source domain distributions.

%Moreoribution.ver,  is computationally efficient and improves TTT performance substantially.  %Contrary to source-free domain adaptation, where information about source domain is often strictly absent, we believe having access to statistical information in the source domain is a realistic assumption for two reasons. First, source domain 
%When source domain information is strictly prohibited, e.g. practitioners only have access to source domain pretrained models, we further differentiate a source-free sTTT setting which might be less common but poses more challenges to the design of TTT algorithms.

In this work, we propose four techniques to enable efficient and accurate sTTT regardless the availability of source domain distribution information. i) We are inspired by the recent progresses in unsupervised domain adaptation~\cite{tang2020unsupervised} that encourages testing samples to form clusters in the feature space. However, separately learning to cluster in the target domain without regularization from source domain does not guarantee improved adaptation~\cite{tang2020unsupervised}. 
To overcome this challenge, we identify clusters in both the source and target domains through a mixture of Gaussians with each component Gaussian corresponding to one category. Provided with the category-wise statistics from source domain as anchors, we match the target domain clusters to the anchors by minimizing the KL-Divergence as the training objective for sTTT. Therefore, we name the proposed method through feature clustering alone as \textit{test-time anchored clustering (TTAC)}. Since test samples are sequentially streamed, we develop an exponential moving averaging strategy to update the target domain cluster statistics to allow gradient-based optimization. ii) Each component Gaussian in the target domain is updated by the test sample features that are assigned to the corresponding category. Thus, incorrect assignments (pseudo labels) will harm the estimation of component Gaussian. To tackle this issue, we are inspired by the correlation between network's stability and confidence and pseudo label accuracy~\cite{lee2013pseudo,sohn2020fixmatch}, and propose to filter out potential incorrect pseudo labels. Component Gaussians are then updated by the samples that have passed the {filterings}. To exploit the filtered out samples, we  incorporate a global feature alignment~\cite{liu2021ttt++} objective. iii) Self-training~(ST) exploits unlabeled data through predicting pseudo labels with existing model and use the predicted pseudo labels as target to further update the model parameters. ST has demonstrated remarkable success for semi-supervised learning~\cite{sohn2020fixmatch} and domain adaptation~\cite{liu2021cycle}, and we hypothesize that self-training on target domain data could benefit TTT as well. However, as we discovered empirically, direct self-training on target domain yields much inferior results compared to anchored clustering. We attribute this phenomenon to the fact that when there is a significant distribution shift between source and target domains pseudo labels predicted on target domain are more likely to be incorrect. As a result, self-training on incorrect pseudo labels leads to inferior accuracy on target domain. To alleviate the impact of distribution shift on self-training, we use anchored clustering to regularize self-training such that we can simultaneously minimize distribution shift and exploit pseudo labels on target domain to update model parameters. We refer to the combined model as \textbf{TTAC++} to acknowledge the importance of anchored clustering. Extensive evaluations have demonstrated the effectiveness of combining anchored clustering and self-training. iv) When source domain distribution information is strictly absent, we propose to infer the source domain distribution by backpropagating classification loss through category-wise distribution parameters. We demonstrate through simple derivation that sampling from the distribution is not necessary during the optimization and the distribution parameters can be learned by efficient gradient descent methods.
%We also demonstrate TTAC is compatible with existing TTT techniques, e.g. contrastive learning branch~\cite{liu2021ttt++}, if source training loss is allowed to be modified. 

We summarize the contributions of this work as below.

\begin{itemize}
\item In light of the confusions within TTT works, we provide a categorization of TTT protocols by {two} key factors. Comparison of TTT methods is now fair within each category.
% \item Based on the categorization we provide the first benchmark comparing existing test time adaptation works under a fair setting.
\item We adopt a realistic TTT setting, namely sTTT. To improve test-time feature learning, we propose TTAC by matching the statistics of the target clusters to the source ones. The target statistics are updated through moving averaging with filtered pseudo labels. 
\item To further exploit the unlabeled target domain data, we incorporate a self-training approach to update model w.r.t. classification loss, and we reveal that regularizing self-training with anchored clustering, referred to as TTAC++, consistently outperforms TTAC with minute additional computation overhead.
\item To enable strict source-free test-time training, we develop a light-weight method to infer source domain distributions. We demonstrate that TTAC++ outperforms state-of-the-art methods under the strict source-free sTTT protocol.
\item The proposed method is demonstrated on five test-time training datasets, among which three datasets~(CIFAR10/100-C \& ImageNet-C) focus on test-time adaptation to corrupted target domains, one~(CIFAR10.1) focuses on selected hard samples and another one~(VisDA) focuses on synthetic-to-real adaptation. We also evaluate test-time training on adversarially attacked target dataset. We demonstrate that TTAC++ achieves the state-of-the-art performance on all benchmarks under multiple TTT protocols.
\end{itemize}

\section{Related Work}

\subsection{Unsupervised Domain Adaptation}
Domain adaptation~\cite{wang2018deep} aims to improve model generalization when source and target data are not drawn i.i.d. Unsupervised domain adaptation~(UDA)~\cite{ganin2015unsupervised,tzeng2014deep,long2015learning} makes an assumption that labeled data is only available in the source domain and target domain data are totally unlabeled. UDA methods often simultaneously learn domain invariant feature representations on both source and target domains to improve generalization. This is achieved by introducing a domain discriminator~\cite{ganin2015unsupervised}, 
Follow-up works improve DA by minimizing a divergence~\cite{gretton2012kernel,sun2016deep,zellinger2017central}, adversarial training~\cite{hoffman2018cycada} or discovering cluster structures in the target data~\cite{tang2020unsupervised}. Apart from formulating DA within a task-specific model, re-weighting has been adopted for domain adaptation by selectively up-weighting conducive samples in the source domain~\cite{jiang2007instance, yan2017mind}. Despite the efforts in UDA, it is inevitable to access the source domain data which may be not accessible due to privacy issues, storage overhead, etc. Deploying DA in more realistic scenarios has inspired research into source-free domain adaptation and test-time training/adaptation.

\subsection{Source-Free Domain Adaptation}
UDA is often implemented by simultaneously updating model parameters on both source and target domain data~\cite{ganin2015unsupervised}. Having access to target domain data during model training may not be practical in real-world applications. For instance, users may buy pretrained model from suppliers and hope to adapt to proprietary data. Access to source domain data could be prohibited due to privacy or data storage issues. Without the access to source data, source-free domain adaptation (SFDA) emerges as a more realistic solution. SFDA is often developed through self-training~\cite{pmlr-v119-liang20a, kundu2020universal,qiu2021source, iwasawa2021test,liang2021source}, self-supervised training~\cite{liu2021ttt++} or clustering in the target domain~\cite{yang2021generalized}. It has been demonstrated that SFDA performs well on seminal domain adaptation datasets even compared against UDA methods~\cite{tang2020unsupervised}. SFDA often requires access to all testing data and model adaptation is carried out by iteratively updating on the testing data. Despite the advantage of not requiring source domain data during model adaptation, the iterative model updating strategy restricts the application of SFDA to scenarios where target domain distribution is fixed and training data in target domain is readily available. In a more realistic DA scenario where data arrives in a stream and inference and adaptation must be implemented simultaneously SFDA will no longer be effective. %Moreover, some statistical information on the source domain does not pose privacy issues and can be exploited to further improve adaptation on target data.

\subsection{Test-Time Training}
Collecting enough samples from target domain and adapt models in an offline manner restricts the application to adapting to static target domain. To allow fast and online adaptation, test-time training (TTT)~\cite{sun2020test,wang2022continual,iwasawa2021test,gandelsman2022test,goyaltest2022,chen2022contrastive,choi2021test} or adaptation (TTA)~\cite{wang2020tent} emerges. TTT tackles a scenario where a distribution shift between source and target domain exists and source model is preferably adapted to target domain in a light-weight fashion. Despite many recent works claiming to be test-time training, we notice a severe confusion over the definition of TTT. In particular, whether training objective must be modified~\cite{sun2020test,liu2021ttt++} and whether sequential inference on target domain data is possible~\cite{wang2020tent,iwasawa2021test}. Therefore, to reflect the key challenges in TTT, we define a setting called sequential test-time training (sTTT) which neither modifies the training objective nor violates  sequential inference. Under the more clear definition, some existing works, e.g. TTT~\cite{sun2020test} and TTT++~\cite{liu2021ttt++} is more likely to be categorized into SFDA. Several existing works~\cite{wang2020tent,iwasawa2021test} can be adapted to the sTTT protocol. Tent~\cite{wang2020tent} proposed to adjust affine parameters in the batchnorm layers to adapt to target domain data. The high TTA efficiency inevitably leads to limited performance gain on the target domain. T3A~\cite{iwasawa2021test} further proposed to update classifier prototype through pseudo labeling. Despite being efficient, updating classifier prototype alone does not affect feature representation for the target domain. Target feature may not form clusters at all when the distribution mismatch between source and target is large enough. In this work we propose to simultaneously cluster on the target domain and match target clusters to source domain classes, namely anchored clustering. To further constrain feature update, we introduce additional global feature alignment and pseudo label filtering. Through the introduced anchored clustering, we achieve test-time training of more network parameters and achieve the state-of-the-art performance.

\subsection{Self-Training}

Training models with predictions from their own has been a long-standing paradigm for learning from unlabeled data. In the realm of semi-supervised learning~\cite{van2020survey}, which aims to exploit few labeled data and large unlabeled, self-training has been widely adopted to produce pseudo labels for unlabeled data. Among these works, label-propagation~\cite{zhu2003semi} is implemented on the deep representations to provide pseudo labels for unlabeled data and self-training is achieved by training with the pseudo labels~\cite{iscen2019label}. FixMatch~\cite{sohn2020fixmatch} utilizes the predictions on weak augmented samples as pseudo label to supervise network training on unlabeled data. MixMatch~\cite{berthelot2019mixmatch} sharpens model prediction to serve as pseudo label for self-training. Self-training recently emerges as a promising solution to domain adaptation by updating model on target pseudo labels~\cite{liu2021cycle,kumar2020understanding,xu2022revisiting}. Some concurrent works also demonstrated that self-training is also effective when source domain data is absent ~\cite{wang2021target,sinha2022test}. In this work, we hypothesize that self-training could benefit test-time training by providing pseudo labels on the target domain samples. More importantly, we discover that self-training alone without any constraint is less effective for TTT under large domain shift due to the high noise in pseudo labels. Through combining anchored clustering and self-training, we demonstrate a significant improvement from the previous state-of-the-art thanks to the improved pseudo label quality.

% \subsection{Out-of-Distribution}

% Building robust test-time training algorithms requires selective adaptation at test stage. In many real applications, e.g. autonomous driving, target domain may contain semantic out of distribution (OOD) samples, which are defined as samples not belonging to the source domain classes. Adapting source domain irrespective of OOD samples may ruin the model's performance on in-distribution samples. The 

% In this section, we briefly review works relevant to learning under out of semantic classes

\section{Methodology}

\begin{figure*}[!htb]
    \centering
    \includegraphics[width=0.99\linewidth]{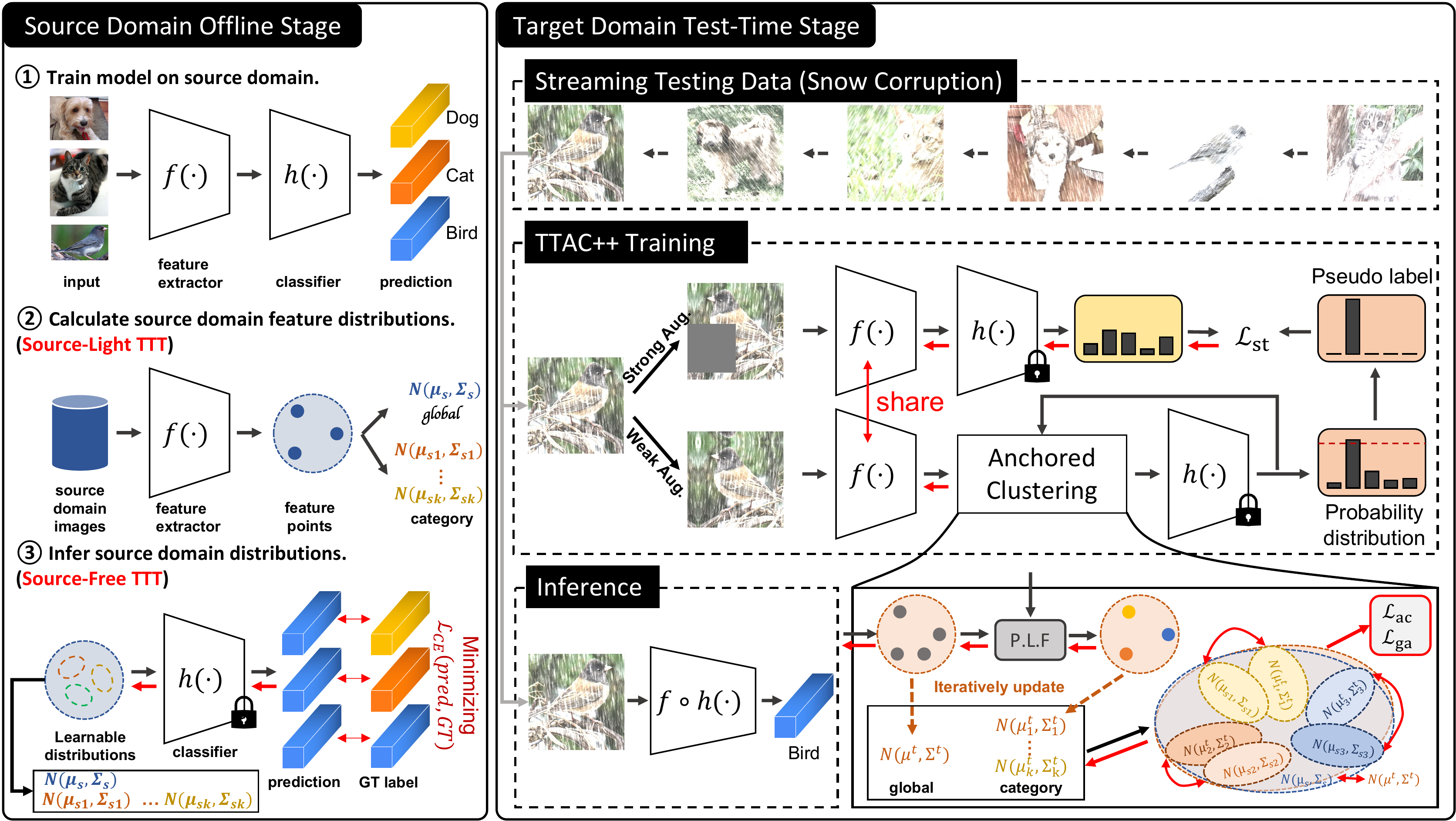}
    \caption{Overview of TTAC++. i) In the source domain offline stage, we calculate or infer category-wise and global distributions in the source domain as anchors. ii) In the test-time stage, testing samples are sequentially streamed and pushed into a fixed-length queue. We apply self-training to testing samples to adapt model weights. Anchored clustering is employed to regularize self-training by aligning source and target domain distributions.}  %for stronger Clusters in target domain are identified through anchored clustering with pseudo label filtering. Target clusters are then matched to the anchors in source domain to achieve test-time training.}
    \label{fig:overview}
\end{figure*}

In this section we first introduce the anchored clustering objective for test-time training through pseudo labeling and then describe an efficient iterative updating strategy. {We then introduce the solution to infer source distribution when source domain data is strictly absent, and self-training for TTT.} For simplicity, We denote the source and target domain datasets as $\set{D}_s=\{x_i,y_i\}_{i=1\cdots N_s}$ and $\set{D}_t=\{x_i\}_{i=1\cdots N_t}$ where a minibatch of target test samples at time stamp $t$ is defined as $\set{B}^t=\{x_i\}_{i=tN_B\cdots (t+1)N_B}$. We further denote the posterior prediction for $x_i$ at time stamp $t$ as $q^t_i=\delta(h(z_i;\vect{w})),\; s.t.\;z_i=f(x_i;\matr{\Theta})$, where $\delta(\cdot)$, $h(\cdot;\vect{w})$ and $f(\cdot;\matr{\Theta})$ denote the a standard softmax function, the classifier head and backbone network, respectively. The $D$ dimensional feature representation is defined as the output of backbone network $z_i=f(x_i;\matr{\Theta})\in\set{R}^{+D}$ due to ReLu activation. An overview of the proposed pipeline is illustrated in Fig.~\ref{fig:overview}.

\subsection{Anchored Clustering for Test-Time Training}

%Feature alignment demonstrated effectiveness for test-time training, however we discover in this work that aligning target testing data feature to source training ones, which is called global feature alignment hereafter, is still suboptimal. First, global feature alignment describes the source and target distributions with two multi-variate Gaussian distributions respectively. In a high dimension feature space with multiple feature clusters, each may correspond to one category, a single-modal Gaussian distribution is insufficient to describe the complex distribution. Moreover, aligning global feature distribution does not guarantee the target data are correctly embedded near the classifier prototype. To tackle these issues, we propose to align feature distribution at a more fine-grained level. Specifically, a mixture of Gaussian is adopted to describe the training and testing feature distributions as,

Inspired by the success of discovering cluster structures in the target domain for unsupervised domain adaptation~\cite{tang2020unsupervised}, we develop an anchored clustering on the test data alone as the initial module for test-time training. %Under the sTTT protocol without the regularization from source labeled data, learningsimultaneously learning cluster center and feature embedding is prone to trivial solution and does not guarantee the reusability of models trained on source data. 
% To tackle this challenge, 
We first use a mixture of Gaussians to model the clusters in the target domain, here each component Gaussian represents one discovered cluster. 
We further use the distributions of each category in the source domain as anchors for the target distribution to match against. In this way, test data features can simultaneously form clusters and the clusters are associated with source domain categories, resulting in improved generalization to target domain. Formally, we first denote the mixture of Gaussians in the source and target domains as,
\begin{equation}
\begin{split}
p_s(z)=\sum_k \alpha_k \mathcal{N}(\mu_{sk},\Sigma_{sk}),\\
p_t(z)=\sum_k \beta_k \mathcal{N}(\mu_{tk},\Sigma_{tk})
\end{split}
\end{equation}
where $\{\mu_k\in\mathbb{R}^d,\Sigma_k\in\mathbb{R}^{d\times d}\}$ represent one cluster in the source/target domain and $d$ is the dimension of feature embedding. Both $\mu_{sk}$ and $\Sigma_{sk}$ can be readily estimated from $\set{D}_s$ through MLE.
Anchored clustering can be achieved by matching the above two distributions and one may directly minimize the KL-Divergence between the two distribution.
%Matching the two distributions without knowing the association between component Gaussians is non-trivial because a good distribution measurement, KL-Divergence, between two mixture of Gaussians does not have 
Nevertheless, this is non-trivial because the KL-Divergence between two mixture of Gaussians has no closed-form solution which prohibits efficient gradient-based optimization. Despite some approximations exist~\cite{hershey2007approximating}, without knowing the semantic labels for each Gaussian component, even a good match between two mixture of Gaussians does not guarantee target clusters are aligned to the correct source ones and this will severely harm the performance of test-time training. %Moreover, fitting a mixture of Gaussian on source data can be achieved in an offline manner, but dynamically fitting a mixture of Gaussian on target data through EM algorithm is subject to erroenous assignment, i.e. data points could be assigned to cluster.
In light of these challenges, we propose a category-wise alignment.  Specifically, we allocate the same number of clusters in both source and target domains, each corresponding to one semantic class, and each target cluster is assigned to one source cluster. We can then minimize the KL-Divergence between each pair of clusters as in Eq.~\ref{eq:KLD}. 
% \vspace{-0.5cm}

\begin{equation}\label{eq:KLD}
\resizebox{0.9\linewidth}{!}{
$
\begin{split}
    \mathcal{L}_{ac}&=\sum_k D_{KL}(\mathcal{N}(\mu_{sk},\Sigma_{sk})||\mathcal{N}(\mu_{tk},\Sigma_{tk}))\\
    &=\sum_k -H(\mathcal{N}(\mu_{sk},\Sigma_{sk})) + H(\mathcal{N}(\mu_{sk},\Sigma_{sk}),\mathcal{N}(\mu_{tk},\Sigma_{tk}))
\end{split}
$
}
\end{equation}

The KL-Divergence can be further decomposed into the entropy $H(\mathcal{N}(\mu_{sk},\Sigma_{sk}))$ and cross-entropy $H(\mathcal{N}(\mu_{sk},\Sigma_{sk}),\mathcal{N}(\mu_{tk},\Sigma_{tk}))$. It is commonly true that the source reference distribution $p_s(z)$ is fixed thus the entropy term is a constant $C$ and only the cross-entropy term is to be optimized.
Given the closed-form solution to the KL-Divergence between two Gaussian distributions, we now write the anchored clustering objective as,

\begin{equation}\label{eq:anchored_clustering_loss}
\resizebox{0.9\linewidth}{!}{
$
\begin{split}
    \mathcal{L}_{ac}= &\sum_k \{\log \sqrt{2\pi^d|\Sigma_{tk}|} + \frac{1}{2}(\mu_{tk}-\mu_{sk})^\top\Sigma_{tk}^{-1}(\mu_{tk}-\mu_{sk})  \\ &+ tr(\Sigma_{tk}^{-1}\Sigma_{sk})\} + C
\end{split}
$
}
\end{equation}

The source cluster parameters, mean and covariance, can be readily estimated in an offline manner by running through the training samples. These information will not cause any privacy leakage and only introduces a small computation and storage overheads. Nevertheless, one might encounter a more constrained scenario where distributional information on the source domain is prohibited, e.g. downstream user can only have access to model architecture and weights. In the following section, we shall introduce a strict source-free anchored clustering method by inferring source domain clusters. To differentiate the settings, we refer to the former one as source-light TTT, where statistical information on source domain is still available, and the latter one as source-free TTT, where no information on source domain is available. %In the next section, we elaborate clustering in the target domain.

% \noindent\textbf{Learning Prototype}

\subsection{Clustering through Pseudo Labeling}\label{sect:cluster_pl}

In order to test-time train network with anchored clustering loss, one must obtain target cluster parameters $\{\mu_{tk},\Sigma_{tk}\}$. 
For a minibatch of target test samples $\set{B}^t=\{x_i\}_{i=tN_B\dots (t+1)N_B}$ at timestamp $t$, %we first denote the predicted posterior as $P^t=softmax(h(f(x_i)))\in[0,1]^{B\times K}$ where $softmax(\cdot)$, $h(\cdot)$ and $f(\cdot)$ respectively denote a standard softmax function, the classifier head and backbone network. 
the pseudo labels are obtained via $\hat{y}_i=\arg\max_k q_{ik}^t$. Given the predicted pseudo labels we could estimate the mean and covariance for each component Gaussian with the pseudo labeled testing samples.
% \begin{equation}\label{eq:naiveupdate}
% \begin{split}
%     \mu_{tk} = \frac{\sum_{i} \mathbbm{1}(\hat{y}_i=k)f(x_i)}{\sum_i \mathbbm{1}(\hat{y}_i=k)},\quad
%     \Sigma_{tk} = \frac{\sum_i \mathbbm{1}(\hat{y}_i=k) (f(x_i)-\mu_{tk})^\top(f(x_i)-\mu_{tk})}{\sum_i\mathbbm{1}(\hat{y}_i=k)}
% \end{split}
% \end{equation}
However, pseudo labels are always subject to model's discrimination ability. The error rate for pseudo labels is often high when the domain shift between source and target is large, directly updating the component Gaussian is subject to erroneous pseudo labels, a.k.a. confirmation bias~\cite{arazo2020pseudo}. To reduce the impact of incorrect pseudo labels, we first adopt a light-weight temporal consistency (TC) pseudo label filtering approach. Compared to co-teaching~\cite{han2018co} or meta-learning~\cite{li2019learning} based methods, this light-weight method does not introduce additional computation overhead and is therefore more suitable for test-time training.
% Specifically, we denote a queue of test time trained models up to a fixed past steps $\{\Theta_t,\Omega_t\}_{t=T-T_w,\cdots T}$, the temporal consistency is defined as,
Specifically, to alleviate the impact from the noisy predictions, we calculate the temporal exponential moving averaging posteriors $\tilde{q}^t \in [0,1]^{N \times K}$ as below,

% \begin{equation}
%     % P^t_i = softmax(h(f(X_i))),\quad 
%     \tilde{P}^t_i = 
%     \left\{
%         \begin{array}{lr}
%             P^t_i & newcomers \\
%             (1 - \alpha) * \tilde{P}^{t-1}_i + \alpha * P^t_i & others
%         \end{array}
%     \right.
% \end{equation}
% \vspace{-0.5cm}

\begin{equation}
    % P^t_i = softmax(h(f(X_i))),\quad 
    \tilde{q}^t_i = 
            (1 - \xi) * \tilde{q}^{t-1}_i + \xi * q^t_i,\quad s.t.\quad\tilde{q}^{0}_i=q^0_i
\end{equation}

The temporal consistency filtering is realized as in Eq.~\ref{eq:tc_filter} where $\tau_{TC}$ is a threshold determining the maximally allowed difference in the most probable prediction over time. If the posterior deviate from historical value too much, it will be excluded from target domain clustering.
% \vspace{-0.2cm}

\begin{equation}\label{eq:tc_filter}
    F_i^{TC} = \mathbbm{1}((q_{i\hat{y}}^t - \tilde{q}^{t-1}_{i\hat{y}}) > \tau_{TC}),\; s.t. \; \hat{y} = \arg\max_k(q_{ik}^t)
\end{equation}

% For those testing samples which do not have a historical predictions, we also incorporate an additional filter based on entropy (\textcolor{red}{is this correct?}) as,

Due to the sequential inference, test samples without enough historical predictions may still pass the TC filtering.  So, we further introduce an additional pseudo label filter directly based on the posterior probability as,
% \vspace{-0.5cm}

\begin{equation}
    F_i^{PP}=\mathbbm{1}(\tilde{q}^{t}_{i\hat{k}}>\tau_{PP})
\end{equation}

By filtering out potential incorrect pseudo labels, we update the component Gaussian only with the leftover target samples as below.
% \vspace{-0.5cm}

\begin{equation}
\centering
\resizebox{0.9\linewidth}{!}{
$
\begin{split}
    &\mu_{tk} = \frac{\sum\limits_{i} F^{TC}_iF^{PP}_i\mathbbm{1}(\hat{y}_i=k)z_i}{\sum\limits_i F^{TC}_iF^{PP}_i\mathbbm{1}(\hat{y}_i=k)},\\
    &\Sigma_{tk} = \frac{\sum\limits_i F^{TC}_iF^{PP}_i\mathbbm{1}(\hat{y}_i=k) (z_i-\mu_{tk})^\top(z_i-\mu_{tk})}{\sum\limits_i F^{TC}_iF^{PP}_i\mathbbm{1}(\hat{y}_i=k)}
\end{split}
$
}
\end{equation}
% \vspace{-0.5cm}

\subsection{Global Feature Alignment}

As discussed above, test samples that do not pass the filtering will not contribute to the estimation of target clusters. Hence, anchored clustering may not reach its full potential without the filtered test samples. To exploit all available test samples, we propose to align global target data distribution to the source one. We approximate the global feature distribution of the source data as $\hat{p}_s(x)=\mathcal{N}(\mu_s,\Sigma_s)$ and the target data as $\hat{p}_t(x)=\mathcal{N}(\mu_t,\Sigma_t)$. To align two distributions, we again minimize the KL-Divergence as,
\vspace{-0.5cm}

\begin{equation}\label{eq:global_loss}
    \mathcal{L}_{ga}=D_{KL}(\hat{p}_s(x)||\hat{p}_t(x))
\end{equation}

Similar idea has appeared in~\cite{liu2021ttt++} which directly matches the moments between source and target domains~\cite{zellinger2017central} by minimizing the F-norm for the mean and covariance, i.e. $||\mu_t-\mu_s||^2_2+||\Sigma_t-\Sigma_s||^2_F$. However, designed for matching complex distributions represented as drawn samples, central moment discrepancy~\cite{zellinger2017central} requires summing infinite central moment discrepancies and the ratios between different order moments are hard to estimate.  For matching two parameterized Gaussian distributions KL-Divergence is more convenient with good explanation from a probabilistic point of view. Finally, we add a small constant to the diagonal of $\Sigma$ for both source and target domains to increase the condition number for better numerical stability.

% This approach is subject to larger norm in the covariance term, due to the higher dimension of $\Sigma\in\mathca{R}^{d\times d}$, and choosing appropriate weight between the two terms would introduce another hyper-parameter. In contrast, minimizing KL-Divergence is hyper-parameter efficient and has better explanation from a probabilistic point of view. 

\subsection{Source-Free TTT by Inferring Source Domain Distributions}\label{sect:SFTTT}

%The proposed anchored clustering approach requires access to light-weight source domain cluster information as anchors. Although the light-weight information does not risk privacy leakage, collecting source domain statistics may not be always feasible when these statistic information are not collected during the training stage. 
In order to enable test-time training under strict source-free setting, we propose to infer the necessary source domain statistical information, i.e. the class-wise mean and covariance matrix, from network weights only. W.o.l.g., we write the classifier head as a linear classifier $h({z}_i;\vect{w})=\vect{w}^\top {z}_i$ by omitting the bias term, which though can still be preserved in a homogeneous coordinate. Without knowing the true class-wise distribution, we hypothesize that each class $k$ is subject to a uni-modal Gaussian distribution $p_{sk}(z)=\mathcal{N}({\mu}_{sk},{\Sigma}_{sk})$ as given in the previous section. Given a model well trained on the source domain we could expect the following class-wise risk being minimized w.r.t. classifier weights.

% \begin{equation}
% \mathcal{L}_{CE}=-\frac{1}{|\set{D}_s|}\sum_{x_i,y_i\in\set{D}_s}\log\sum_k[y_i=k]\sigma(w_kz_i)
% \end{equation}
\begin{equation}
\begin{split}
    \mathcal{L}_{sk}(\vect{w},\matr{\Theta})=&\mathbbm{E}_{z\sim p_{sk}(z)}[-\log\delta(\vect{w}_k^\top z)]\\
    =&\mathbbm{E}_{\tilde{z}\sim \mathcal{N}(\vect{0},\vect{I})}[-\log\delta(\vect{w}_k^\top(\mu_{sk}+\matr{A}_{sk}\tilde{z}))]
\end{split}
\end{equation}
where $\matr{A}_{sk}\matr{A}_{sk}^\top=\matr{\Sigma}_{sk}$ satisfies a Cholesky decomposition. When source domain feature distribution is unknown while the classifier head $\vect{w}$ is available, we could rewrite the above optimization by substituting the optimization variables to source domain class-wise distribution as below, where the lower bound is derived according to Jensen inequality, as $-\log\delta(\cdot)$ is a convex function. The equality holds when $\matr{A}_{sk}=\matr{0}$.
% \begin{equation}\label{eq:learnproto}
\begin{align}
    \hat{\mathcal{L}}_{sk}(\mu_{sk},\matr{A}_{sk})=&\mathbbm{E}_{\tilde{z}\sim \mathcal{N}(\vect{0},\vect{I})}[-\log\delta(\vect{w}_k^\top(\mu_{sk}+\matr{A}_{sk}\tilde{z}))]\label{eq:learnproto}\\
    \geq&-\log\delta(\mathbbm{E}_{\tilde{z}\sim\mathcal{N}(\matr{0},\matr{I})}[\vect{w}_k^\top(\mu_{sk}+\matr{A}_{sk}\tilde{z})])\\
    =&-\log\delta(\vect{w}_k^\top\vect{\mu}_{sk})\label{eq:lowerbound}
\end{align}
% \end{equation}

% \begin{equation}
% \resizebox{0.9\linewidth}{!}{$
% \begin{aligned}
% \hat{\mathcal{L}}{sk}(\mu{sk},\matr{A}{sk})=&\mathbbm{E}{\tilde{z}\sim \mathcal{N}(\vect{0},\vect{I})}[-\log\delta(\vect{w}k^\top(\mu{sk}+\matr{A}{sk}\tilde{z}))]\label{eq:learnproto}\\
% \geq&-\log\delta(\mathbbm{E}{\tilde{z}\sim\mathcal{N}(\matr{0},\matr{I})}[\vect{w}k^\top(\mu{sk}+\matr{A}_{sk}\tilde{z})])\\
% =&-\log\delta(\vect{w}k^\top\vect{\mu}{sk})\label{eq:lowerbound}
% \end{aligned}
% $}
% \end{equation}

We interpret this problem as discovering the source domain class-wise distribution such that samples drawn from these distributions can be correctly classified. We argue that directly optimizing Eq.~\ref{eq:learnproto} without any constraint on $\matr{A}_{sk}$ is equivalent to optimizing the lower bound, Eq.~\ref{eq:lowerbound}. Because any non-zero $\matr{A}_{sk}$ enables the inequality and without constraining $\matr{A}_{sk}$, a trivial solution with $\matr{A}_{sk}=\matr{0}$ exists. %i.e. $\matr{A}_{sk}=\matr{0}$, because any non-zero $\matr{A}_{sk}$ enables the inequality. %, we defer the proof to the Appendix. 
Alternatively, one could fix the covariance matrix, $\matr{\Sigma}_{sk}$, and only update class-wise mean, $\vect{\mu}_{sk}$, and this requires Monte Carlo sampling from a standard multi-variate Gaussian distribution should Eq.~\ref{eq:learnproto} be the objective to optimize. To get rid of the excessive computation of sampling, we empirically figure out an efficient way to infer source-domain distributions by fixing $\matr{\Sigma}_{sk}=\gamma\matr{I}$ and optimizing the lower bound, Eq.~\ref{eq:lowerbound}, to estimate $\vect{\mu}_{sk}$. Moreover, since all backbone features $z_i$ are positive due to ReLu activation, $\mu$ should be all positive so that it may overlap with the true distribution of source domain features. For this purpose, we parameterize $\mu=\hat{\mu}^2$ where $\hat{\mu}$ is unconstrained, and add weight decay to $\hat{\mu}$ to limit the norm of $\mu$.%We choose a is chosen to be 
%For distribution alignment, both category-wise and global mean and covariance are necessary information. 
%We Given the feature representations are taken from the last layer of backbone network after ReLu activation, the 
% Given a inferred source-domain distribution which can be correctly by the source domain classifier, $\mathcal{N}(\tilde{\mu}_{sk},\tilde{\Sigma}_{sk})$, multiplying the distribution mean with an arbitrary scale will further decrease the risk~(cross entropy loss) in Eq.~\ref{eq:learnproto}.%, we defer the proof to the Appendix. 
%To avoid estimating overly large norm for $\mu_{sk}$, we add a weight decay for $\mu_{sk}$. %Considering that the norm of estimated mean may differ from source domainIn anchored clustering, we normalize Normalize norm
%$z_i = \frac{z_i||\mu_k||}{||z_i||}$

The global feature distribution is approximated by a uni-modal Gaussian distribution. Therefore, to infer the global feature distribution, we  use a single Gaussian distribution $\mathcal{N}(\mu_s,\Sigma_s)$ to approximate the mixture of per-category Gaussians. Specifically, the following KL-Divergence is minimized with a closed-form solution~\cite{hershey2007approximating}. 

\begin{equation}
\centering
\resizebox{0.89\linewidth}{!}{
$
\begin{split}
    &\mu_{s}^*,\Sigma_{s}^* = \arg\min_{\mu_{s},\Sigma_{s}} D_{KL}(\mathcal{N}(\mu_s,\Sigma_s)||\sum_k\frac{1}{K}\mathcal{N}(\mu_{sk},\Sigma_{sk}))\\
    &\Rightarrow\mu_s^* = \sum_k\frac{1}{K} \mu_{sk}\\
    &\Rightarrow\Sigma_s^* = \sum_k \frac{1}{K}(\Sigma_{sk}+(\mu_{sk}-\mu_s)(\mu_{sk}-\mu_s)^\top) %= \Sigma_{sk} + \Sigma_{\mu sk}
\end{split}
$
}
\end{equation}
%where we denote $\sum_k \frac{1}{K}(\mu_{sk}-\mu_s)(\mu_{sk}-\mu_s)^\top$ as $\Sigma_{\mu sk}$ for the covariance of all class-wise means $\mu_{sk}$.

\subsection{Efficient Iterative Updating}

% We choose to model the feature distribution via a single multi-variate Gaussian distribution for the simplicity of Cross-Entropy loss. 

Despite the distribution for source data can be trivially estimated from all available training data in a totally offline manner, estimating the distribution for target domain data is not equally trivial, in particular under the sTTT protocol.
In a related research~\cite{liu2021ttt++}, a dynamic queue of test data features are preserved to dynamically estimate the statistics, which will introduce additional memory footprint~\cite{liu2021ttt++}. %Moreover, as the gradient only exists for the current mini-batch, a long queue would vanish the learning gradient, i.e. $1/N_q$-th the original loss where $N_q$ is the length of queue.
To alleviate the memory cost we propose to iteratively update the running statistics for Gaussian distribution. 
%Formally, we define $t$-th test minibatch as $\set{B}^t=\{x_i\}_{i=1\cdots N_{B}}$. 
Denoting the running mean and covariance at time stamp $t$ as $\mu^t$ and $\Sigma^t$, we present the rules to update the mean and covariance in Eq.~\ref{eq:runningstatistics}. More detailed derivations and update rules for per cluster statistics are deferred to the Supplementary. 

% \vspace{-0.5cm}
\begin{equation}\label{eq:runningstatistics}
    \resizebox{0.89\linewidth}{!}{
    $
\begin{split}
% &\mu^t = \mu^{t-1}+\sum_{x_i\in\set{B}}\alpha_i(f(x_i)-\mu^{t-1}), \\
    % &N^t = N^{t-1} + |\set{B}^t|,\quad
    % \delta^t=\frac{1}{N^t}{\sum\limits_{x_i\in\set{B}}(f(x_{i}) - \mu^{t-1})},\\
    % &\Sigma^t=\Sigma^{t-1}+\sum_{x_i\in\set{B}}\alpha_i^2(f(x_i)-\mu^{t-1})^\top(f(x_i)-\mu^{t-1})-\Sigma^{t-1} - \delta^{t-1}^\top\delta^t
    % &\sigma_{mn}[t]=\sigma_{mn}[t-1] + \frac{\sum\limits_{i=N_{t-1}}^{N_t}([f_m(X_i)_{i} - \mu_m[t-1]][f_n(X_i) - \mu_n[t-1]] - \sigma_{mn}[t-1])}{N_t} - \delta_m[t]\cdot\delta_n[t]
    %
    & \mu^t = \mu^{t-1} + \delta^t, \\
    &\Sigma^t=\Sigma^{t-1}+a^t{\sum_{x_i\in\set{B}}\{(z_i-\mu^{t-1})^\top(z_i-\mu^{t-1})-\Sigma^{t-1}\}} \\
    &- {\delta^t}^\top\delta^t \\
    & \delta^t=a^t{\sum\limits_{x_i\in\set{B}}(z_i - \mu^{t-1})},\quad 
    N^t = N^{t-1} + |\set{B}^t|, \quad 
    a^t = \frac{1}{N^t}, 
\end{split}
$
}
\end{equation}

% Furthermore, the running mean and covariance for the $k^{th}$ component Gaussian are denoted as $\mu_k^t$ and $\Sigma_k^t$ respectively. Eq.~\ref{eq:category-wise-runningstatistics} are the rules to update them.

% \begin{equation}\label{eq:category-wise-runningstatistics}
% \begin{split}
%     & \delta_k^t=a_k^t{\sum_{x_i\in\set{B}}F^{TC}_iF^{PP}_i\mathbbm{1}(\hat{y}_i=k)(f(x_i)-\mu_k^{t-1})}, \\
%     & N_k^t = N_k^{t-1} + \sum_{x_i\in\set{B}}F^{TC}_iF^{PP}_i\mathbbm{1}(\hat{y}_i=k), \quad 
%     a_k^t = \frac{1}{N_k^t},\\
%     & \mu_k^t = \mu_k^{t-1} + \delta_k^t, \\
%     &\Sigma_k^t=\Sigma_k^{t-1}+a_k^t{\sum_{x_i\in\set{B}}\{F^{TC}_iF^{PP}_i\mathbbm{1}(\hat{y}_i=k)(f(x_i)-\mu_k^{t-1})^\top(f(x_i)-\mu_k^{t-1})-\Sigma_k^{t-1}\}} - {\delta_k^t}^\top\delta_k^t
% \end{split}
% \end{equation}

Since $N^t$ grows larger overtime, new test samples will have smaller contribution to the update of target domain statistics when $N^t$ is large enough. As a result, the gradient calculated from current minibatch will vanish. To alleviate this issue, we impose a clip on the value of $\alpha^t$ as below. As such, the gradient can maintain a minimal scale even if $N^t$ is very large. 

% change the coefficients $a_k^t$ and $a^t$ to constants while they are less than thresholds as,

% New-come test samples thus obtain few weight of updating the corresponding mean and covariance, since $a^t$ and $a_k^t$ are too small. It's the situation that the new features generated by the adapted model have the same weight and the relatively incorrect old features generated by the pre-adapted model, which impacts the estimation of the real time distribution of target domain feature. To alleviate this situation, we change the coefficients $a_k^t$ and $a^t$ to constants while they are less than thresholds as,
% \vspace{-0.3cm}

\begin{equation}
% \begin{split}
    a^t = \left \{
        \begin{array}{lcl}
            \frac{1}{N^t} & & N^t < N_{clip} \\
            \frac{1}{N_{clip}} & & others
        \end{array}
        \right.
    % a_k^t = \left \{
    %     \begin{array}{lcl}
    %         \frac{1}{N_k^t} & & N_k^t < n_{category} \\
    %         \frac{1}{n_{categroy}} & & others
    %     \end{array}
    %     \right.
% \end{split}
\end{equation}

\subsection{Self-Training for TTT}

%Matching the distribution between source and target domains regularizes the network to extract features on target domain to reuse the source domain classifiers. However, optimizing the distribution matching loss alone does not impose a strong constraint on the correctness of classification. Therefore, we propose to adopt self-training~(ST) on the test data to impose stronger test-time adaptation. 
Self-training~(ST) has been widely adopted in semi-supervised learning where predictions on unlabeled data are admitted as pseudo labels, and model is trained with the pseudo labels~\cite{berthelot2019mixmatch,sohn2020fixmatch}. In this work, we explore employing self-training for TTT. Blindly taking all pseudo labels for training has been demonstrated to deteriorate the performance as incorrect pseudo labels act as noisy labels and a high percentage of noisy label is harmful for model training. This phenomenon is also referred to as confirmation bias~\cite{arazo2020pseudo}. As demonstrated in the empirical evaluations in Sect.~\ref{sect:ttt_eval}, self-training alone is not guaranteed to outperform competing methods. The performance may even degrades after observing enough testing samples as shown in Fig.~\ref{fig:imagenet_cumulative}.
To reduce the impact of confirmation bias, we first propose to employ anchored clustering as regularization. As anchored clustering allows better alignment between source and target feature distributions, self-training is able to benefit from more accurate pseudo labels and the model is less likely to be harmed by the wrong pseudo labels. This can be achieved by simultaneously optimizing anchored clustering losses and self-training loss. In addition, we further take an approach similar to \cite{sohn2020fixmatch} by filtering out less confident pseudo labels for self-training as in Eq.~\ref{eq:selftraining}. 

% In this work, we adopt a The self-training module uses the pseudo label predicted on target domain testing sampler and utilizes it to supervise network's prediction with cross-entropy loss.

% With the helps of anchored clustering and global feature alignment, to some certain extend, the feature distribution in target domain has been regularized to one in source domain. However, when two distributions are relatively close, the constraints of above components are very weak, especially for each sample. To further enhance the robustness and effectiveness of the test-time adaptation model, we propose the self-training module. The self-training module is consisted of the consistency regularization proposed in FixMatch~\cite{sohn2020fixmatch}, which generates an artificial label by the weakly-augmented version of a given test sample and utilizes it to supervises the strongly-augmented version of the same sample with the cross-entropy loss. Instead of a supervised loss used for labeled data in FixMatch, we leverage the anchored clustering and global feature alignment modules to alleviate the feature collapse during test-time. We formulas the consistency regularization as below,

\begin{equation}\label{eq:selftraining}
\resizebox{0.91\linewidth}{!}{
$
    \mathcal{L}_{st} = \frac{1}{|\set{B}_t|}\sum\limits_{x_i\in\set{B}_t}{\mathbbm{1}(\max\limits_k(q_k(\mathcal{W}(x_i))) \ge \tau_{st})H(\hat{y}_i, q(\mathcal{A}(x_i)))}
$
}
\end{equation}
where $q(\cdot) = \sigma(h(f(\cdot)))$ denotes the probabilistic posterior, $\hat{y}_i = \arg\max_k(q(\mathcal{W}(x_i)))$ denotes the predicted pseudo label, $\mathcal{W}(\cdot)$ denotes a weak augmentation operation consisting of RandomHorizontalFlip and RandomResizedCrop, $\mathcal{A}(\cdot)$ denotes a strong augmentation operation implemented as RandAugment~\cite{RandAugment}, and $\tau_{cr}$ denotes the confidence threshold.

\subsection{TTAC++ Training Algorithm}
We summarize the training algorithm for the TTAC++ in Algo.~\ref{alg:main}. For effective clustering in target domain, we allocate a fixed length memory space, denoted as testing time queue $\set{C} \in \set{R}^{N_{C} \times H \times W \times \texttt{3}}$, to store the recent testing samples. In the sTTT protocol, we first make instant prediction on each testing sample, and only update the model when $N_B$ testing samples are accumulated. TTAC++ can be efficiently implemented, e.g. with two devices, one is for continuous inference and another is for model updating.%The number of the iterations performed in each adaptation stage is denoted as $n$.

% \vspace{-0.4cm}
\begin{algorithm}
% \setstretch{1.2}
\caption{TTAC++ Algorithm }\label{alg:main}
% \begin{algorithmic}
% \Require $n \geq 0$
% \Ensure $y = x^n$
% \State $y \gets 1$
% \State $X \gets x$
% \State $N \gets n$
% \While{$N \neq 0$}
% \If{$N$ is even}
%     \State $X \gets X \times X$
%     \State $N \gets \frac{N}{2}$  \Comment{This is a comment}
% \ElsIf{$N$ is odd}
%     \State $y \gets y \times X$
%     \State $N \gets N - 1$
% \EndIf
% \EndWhile
\SetKwInOut{Input}{input}
\SetKwInOut{Return}{return}
\Input{A new testing sample batch $\set{B}^t=\{x_i\}_{i=tN_B\dots (t+1)N_B}$.}

\textcolor{gray}{\# Update the testing sample queue $\set{C}$.}

$\set{C}^t=\set{C}^{t-1}\setminus \set{B}^{t-N_C/N_B}$,\quad
$\set{C}^t=\set{C}^t\bigcup \set{B}^{t}$

\For{$ 1 $ \KwTo $N_{itr}$}
{
    \For{minibatch $\{x^t_i\}^N_{i=1}$ in  $\set{C}^t$}
    {
        \textcolor{gray}{\# Generate weak and strong augmented samples}
        
         $\mathcal{W}(x^t_i), \quad \mathcal{A}(x^t_i)$

        \textcolor{gray}{\# Obtain the predicted posterior and pseudo labels}
        
        $q(\mathcal{W}(x_i^t))=\sigma(h(f(\mathcal{W}(x_i^t))))$
        
        $\hat{y}_i=\arg\max_k(q(\mathcal{W}(x_i^t)))$
        
        \textcolor{gray}{\# Update the global and per-cluster running mean and covariance by Eq.~\ref{eq:runningstatistics} with $\mathcal{W}({x_i^t})$}
        
        $\mu^t$,\quad $\Sigma^t$,\quad $\{\mu_k^t\}$,\quad $\{\Sigma_k^t\}$
        
        % $\Sigma^t=\Sigma^{t-1}+a^t{\sum_{x^t_i\in\set{B}^{t}}\{(f^t_i-\mu^{t-1})^\top(f^t_i-\mu^{t-1})-\Sigma^{t-1}\}} - {\delta^t}^\top\delta^t$
        
        % \textcolor{gray}{\# calculate the running mean and covariance of each cluster (in Eq.~\ref{eq:category-wise-runningstatistics})}
        
        % $\mu_k^t = \mu_k^{t-1} + \delta_k^t$
        
        % $\Sigma_k^t=\Sigma_k^{t-1}+a_k^t{\sum_{x_i\in\set{B}}\{F^{TC}_iF^{PP}_i\mathbbm{1}(\hat{y}_i=k)(f^t_i-\mu_k^{t-1})^\top(f^t_i-\mu_k^{t-1})-\Sigma_k^{t-1}\}} - {\delta_k^t}^\top\delta_k^t$
        
        \textcolor{gray}{\# Calculate the anchored clustering losses according to Eq.~\ref{eq:anchored_clustering_loss} and Eq.~\ref{eq:global_loss}}
        
        $\mathcal{L}_{ac}+\lambda_1\mathcal{L}_{ga}$

        \textcolor{gray}{\# Calculate self-training loss according to  Eq.~\ref{eq:selftraining}}
        
        $\mathcal{L}_{st}$%$ = \frac{1}{N_B}\sum_{i=1}^{N_B}{\mathbbm{1}(max(q(a^t_i)) \ge \tau_{cr})H(\hat{q}(a^t_i), q(b^t_i))}$

       \textcolor{gray}{\# One step gradient descent on the total loss}
       
       $\mathcal{L}_{ac} + \lambda_1\mathcal{L}_{ga}+ \lambda_2\mathcal{L}_{st}$
        
        %update network $f$ to minimize $\mathcal{L}$
    }
}

% \Return{backbone network $f(\cdot)$}
% \Return{the prediction result of the latest test sample $argmax_k(h(f(\set{B}^t)))$}

% \end{algorithmic}
\end{algorithm}
% \vspace{-0.5cm}

% \subsection{Additional Feature Alignment}

% The proposed anchored clustering aims to match the clusters discovered in the target domain to the source domain clusters so that the classifiers can be reused. This strategy is mainly effective when enough target domain data are observed and may not perform well at the beginning of test-time training. To improve the TTT performance at the early stage, we can further align the global feature distributions to enhance the feature alignment. The objective is similar to the feature alignment proposed in~\cite{liu2021ttt++}, but we adopt the KL-Divergence between source and target global feature distributions as the objective to optimize as below.

% \begin{equation}
% H(P_s,P_t)= \log \sqrt{2\pi^d|\Sigma_t|} + \frac{1}{2}(\mu_t-\mu_s)^\top\Sigma_t^{-1}(\mu_t-\mu_s) + tr(\Sigma_t^{-1}\Sigma_s)
% \end{equation}

% The above Cross-Entropy objective is better than directly minimizing the F-norm between the mean and covariance, i.e. $||\mu_t-\mu_s||^2_2+||\Sigma_t-\Sigma_s||^2_F$, as adopted by ~\cite{liu2021ttt++} in that the mean and covariance are properly weighted in accordance with distribution matching while direct minimizing the F-norm is subject to larger norm in the covariance term and choosing appropriate weight between the two terms would introduce another hyper-parameter. The advantage of using Cross-Entropy loss is also demonstrated via empirical evaluation on multiple test-time training datasets.

\section{Experiment}

% In this section, we first demonstrate the differences on the settings between source-free domain adaptation and test-time training. 
% Source-free domain adaptation setting is an offline adaptation setting, on which these approaches\cite{pmlr-v119-liang20a, liu2021ttt++} are allowed to adapt their model on the target dataset by multiple epoches. Yet test-time training setting is an online adaptation setting, which allows these approaches\cite{ioffe2015batch, sun2020test, wang2020tent, iwasawa2021test} 

In the experiment section, we first compare against existing methods on different test-time training protocols based on the two key factors. We then ablate the effectiveness of each component in TTAC++. Further analysis on the cumulative performance, qualitative insights, etc. are provided at the end of this section. 

% \vspace{-0.2cm}
\subsection{Datasets}
\begin{figure}
    \centering
    \resizebox{1.0\linewidth}{!}{
        \includegraphics{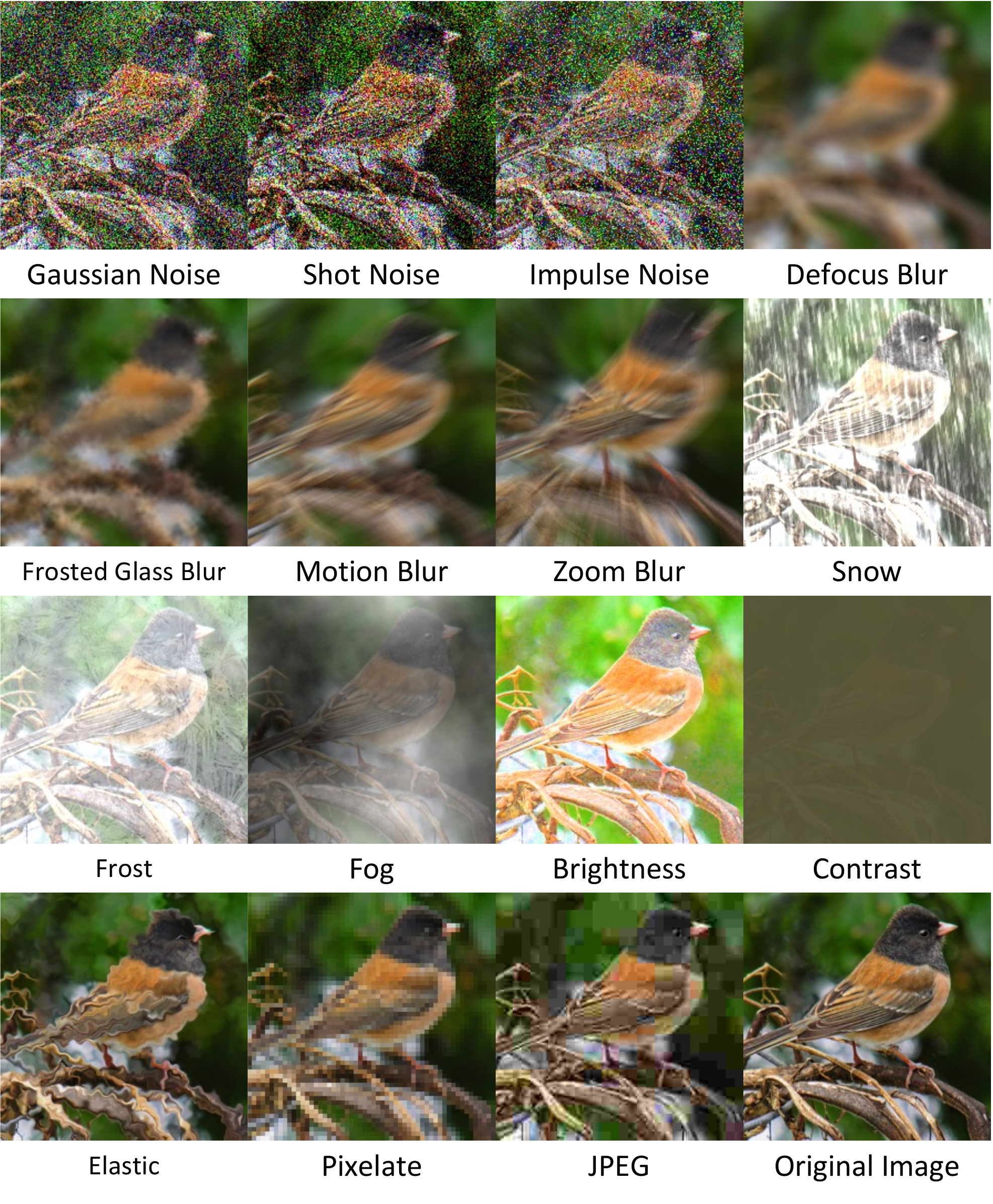}
    }
    \caption{Illustration of corruptions on target domain images. Examples are selected from the ImageNet-C dataset.}
    \label{fig:corruptedimages}
\end{figure}

We evaluate on five test-time training datasets and report the classification error rate~($\%$) throughout the experiment section. To evaluate the test-time training efficacy on corrupted target images, we use \textbf{CIFAR10-C/CIFAR100-C}~\cite{hendrycks2018benchmarking}, each consisting of 10/100 classes with 50,000 training samples of clean data and 10,000 corrupted test samples.   %\textcolor{red}{Additionally, for the large-scale experiments, we choose \textbf{ImageNet-C}~\cite{ILSVRC15}, which has 1,000 classes and 15 types of corruption test set generated with 50,000 samples from original evaluation set.}
We further evaluate test-time training on hard target domain samples with \textbf{CIFAR10.1}~\cite{pmlr-v97-recht19a}, which contains around 2,000 difficult testing images sampled over years of research on the original CIFAR-10 dataset.
%To evaluate the adaptation capability facing the normal domain shift, \textbf{CIFAR10.1}~\cite{pmlr-v97-recht19a}, which contains roughly 2,000 new test images that were sampled after multiple years of research on the original CIFAR-10 dataset, is chosen.
To demonstrate the ability to do test-time training for synthetic data to real data transfer we further use \textbf{VisDA-C}~\cite{VisDA}, which is a challenging large-scale synthetic-to-real object classification dataset, consisting of 12 classes, 152,397 synthetic training images and 55,388 real testing images. To evaluate large-scale test-time training, we use \textbf{ImageNet-C}~\cite{hendrycks2018benchmarking} which consists of 1,000 classes and 15 types of corruptions on the 50,000 testing samples. Some qualitative examples of common corruptions on ImageNet-C are illustrated in Fig.~\ref{fig:corruptedimages}.
% Finally, to evaluate test-time training on 3D point cloud data, we choose \textbf{ModelNet40-C}~\cite{ModelNet40-C}, which consists of 15 common and realistic corruptions of point cloud data, with 9,843 training samples and 2,468 test samples.
Finally, we also evaluate the effectiveness of test-time training against adversarial attacks by implementing TTT on generated \textbf{adversarial samples} on CIFAR-10 dataset, which is referred to as CIFAR-10-adv throughout this work. 

% \vspace{-0.2cm}
\subsection{Experiment Settings}

\subsubsection{Hyperparameters}
We use the ResNet-50~\cite{he2016deep} as backbone network for fair comparison with existing methods. In addition, ViT~\cite{dosovitskiy2020vit} was adopted as backbone for evaluation on the compatibility with transformer architectures.
%for image datasets and the DGCNN~\cite{wang2019dynamic} on ModelNet40-C. 
We train the backbone network $f(\cdot)$ by SGD optimizer with momentum on all datasets. On CIFAR10-C/CIFAR100-C and CIFAR10.1, we set the batchsize to 256 and the learning rate to 0.01, 0.0001 and 0.01 respectively. On VisDA-C we set the batchsize to 128 and the learning rate to 0.0001. Hyperparameters are shared across multiple TTT protocols except for $N_C$ and $N_{itr}$ which are only applicable under one-pass adaptation protocols. 
 $\alpha_k$ and $\beta_k$ respectively represent the prevalence of each category, here we set them to 1 over the number of categories. $N_C$ indicates the length of the testing sample queue $C$ under the sTTT protocol, and $N_{itr}$ controls the update epochs on this queue. $\tau_{TC}$ and $\tau_{PP}$ are the thresholds used for  pseudo label filtering. $N_{clip}$ and $N_{clip\_k}$ are the upper bounds of sample counts in the iterative updating of global statistics and target cluster statistics respectively. Finally $\lambda_1$ and $\lambda_2$ are the coefficients of $\mathcal{L}_{ga}$ and $\mathcal{L}_{cr}$ respectively, which are 1 and 10 respectively. The details of each individual hyperparamter are found in Table.~\ref{tab:hyperparameters}. When source domain distribution information is not available, we estimate source domain distributions by minimizing Eq.~\ref{eq:learnproto} with RMSprop optimizer, learning rate 0.001 and weight decay 0.001. We choose $\gamma=max(svdvals(\Sigma_{\mu sk}))/30.$ for the fixed covariance matrix. Wall-Clock times for competing methods are recorded under PyTorch 1.10.2 framework, CUDA 11.3 and a single NVIDIA RTX 3090 GPU.

\begin{table}[]
    \centering
    \caption{Hyper-parameters used on different datasets.}
          \setlength\tabcolsep{4pt} % default value: 6pt
    \resizebox{\linewidth}{!}{
        \begin{tabular}{l|cccccccccccc}
        \toprule
          Dataset & $\alpha_k$ & $\beta_k$ & $N_C$ & $N_{itr}$ & $\xi$ & $\tau_{TC}$ & $\tau_{PP}$ & $\tau_{cr}$ & $N_{clip}$ & $N_{clip\_k}$ & $\lambda_1$ & $\lambda_2$\\
        \midrule
          CIFAR10-C  &  0.1 & 0.1 & 4096 & 4 & 0.9 & 0.95 & -0.001 & 0.9 & 1280 & 128 & 1.0 & 10.0\\
          CIFAR100-C  & 0.01  & 0.01 & 4096 & 4 & 0.9 & 0.95 & -0.001 & 0.9 & 1280 & 64 & 1.0 & 10.0\\
          CIFAR10.1 & 0.1 & 0.1 & 4096 & 4 & 0.9 & 0.95 & -0.001 & 0.9 & 1280 & 128 & 1.0 & 10.0\\
          VisDA-C & $\frac{1}{12}$ & $\frac{1}{12}$ & 4096 & 4 & 0.9 & 0.95 & -0.01 & 0.9 & 1536 & 128 & 1.0 & 10.0\\
          % ModelNet40-C & 0.025 & 0.025 & 4096 & 6 & 0.9 & -0.1 & 0.5 & 1280 & 128 & 1.0\\
          ImageNet-C & 0.001 & 0.001 & 4096 & 2 & 0.9 & 0.95 & -0.01 & 0.9 & 1280 & 64 & 1.0 & 10.0\\
        \bottomrule
        \end{tabular}
    }
    \label{tab:hyperparameters}
\end{table}

\subsubsection{Test-Time Training Protocols}
We categorize test-time training protocols based on two key factors. First, whether the training objective must be changed during training on the source domain, we use Y and N to indicate if training objective is allowed to be changed or not respectively. Second, whether testing data is sequentially streamed and predicted, we use O to indicate a sequential \textbf{O}ne-pass inference and M to indicate non-sequential inference, a.k.a. \textbf{M}ulti-pass inference. With the above criteria, we summarize 4 test-time training protocols, namely N-O, Y-O, N-M and Y-M, and the strength of the assumption increases from the first to the last protocols. 
Ours sTTT setting  makes the weakest assumption, i.e. N-O. Existing methods are categorized by the four TTT protocols, we notice that some methods can operate under multiple protocols.

\noindent\textbf{Source-Free Test-Time Training}.
The proposed TTAC++ relies on aligning the source and target domain distributions. It is often realistic to assume having access to the source domain data distributions, which are light-weight and there is no risk of privacy leakage. Nevertheless, for a fair comparison with existing methods that are strictly \textbf{source-free}, we adopt inferring the source domain statistics from source domain model weights only as introduced in Section~\ref{sect:SFTTT}. Therefore, we differentiate the source-free~(SF) approach from the source-light~(SL) approach in the TTT evaluation protocol. We summarize all protocols evaluated in this work in Table.~\ref{tab:TTTProtocols}.

\begin{table}[!htb]
    \centering
    \caption{The used components under different TTT protocols.}
    \resizebox{0.95\linewidth}{!}{
        \begin{tabular}{c|ccc}
        \toprule
        Protocol & Source Domain Statistics & Contrastive Branch & Multiple Passes \\
        \midrule
        N-O-SF & - & - & - \\
        N-O-SL & \checkmark & - & - \\
        Y-O-SL & \checkmark & \checkmark & - \\
        N-M-SF & - & - & \checkmark \\
        N-M-SL & \checkmark & - & \checkmark \\
        Y-M-SL & \checkmark & \checkmark & \checkmark \\
        \bottomrule
        \end{tabular}
    }
    \label{tab:TTTProtocols}
\end{table}

\begin{table*}[!htb]
    \centering
    %\caption{Different Methods are performed under different Domain Adaptation Settings (e.g. N represents no need to modify network, Y represents need to modify network, O represents one pass to adapt and M represents multiple passes to adapt) and different benchmark datasets (e.g. CIFAR10-C, CIFAR100-C, ModelNet40-C). Also, we give out the assumption strengths for different settings. From the below table, our model performs the state-of-the-art under whichever DA Setting    and whichever dataset.}
   \caption{Comparison under different TTT protocols. Y/N indicates modifying source domain training objective or not. O/M indicate one pass or multiple passes test-time training. SF/SL indicate source-free and source-light respectively. C10-C, C100-C and C10.1 refer to CIFAR10-C, CIFAR100-C and CIFAR10.1 datasets respectively. All numbers indicate error rate in percentage ($\%$).}
    \resizebox{0.5\linewidth}{!}{
        \begin{tabular}{l|cc|ccc}
        \toprule
        Method  & TTT Protocol & Assum. Strength & C10-C & C100-C & C10.1 \\
        \midrule
        TEST &  - & - & 29.15 & 60.34 & 12.10  \\
        \midrule
        BN~\cite{ioffe2015batch} &  N-O-SF & Weak & 15.49 & 43.38 & 14.00 \\
        TENT~\cite{wang2020tent} &  N-O-SF & Weak & 14.27 & 40.72 & 14.40 \\
        T3A~\cite{iwasawa2021test} &  N-O-SF & Weak & 15.44 & 42.72 & 13.50 \\
        SHOT~\cite{pmlr-v119-liang20a} &  N-O-SF & Weak & 13.95 & 39.10 & 13.70 \\
        Conjugate PL~\cite{goyal2022test} & N-O-SF & Weak & 13.21 & 39.39 & 14.20\\
        Self-Training~\cite{sohn2020fixmatch} &  N-O-SF & Weak & 14.66 & 39.25 & \textbf{12.85}\\
        TTAC++~(Ours) & N-O-SF & Weak & \textbf{11.62} & \textbf{37.76} & \textbf{12.85}\\
        \midrule
        TTT++~\cite{liu2021ttt++} &  N-O-SL & Weak & 13.69 & 40.32 & 13.65 \\
        TTAC~\cite{su2022revisiting}  &  N-O-SL & Weak & 10.94 & 36.64 & 12.80 \\
        TTAC+SHOT~\cite{su2022revisiting} & N-O-SL & Weak & 10.99  & 36.39 & 12.40 \\
        TTAC++~(Ours) & N-O-SL & Weak & \textbf{9.78} & \textbf{35.48} & \textbf{12.20} \\
        \midrule
        TTT++~\cite{liu2021ttt++} & Y-O-SL & Medium & 13.00 & 35.23 & 12.60 \\
        TTAC~\cite{su2022revisiting} & Y-O-SL & Medium & 10.69 & 34.82 & 12.00 \\
        TTAC++~(Ours) & Y-O-SL & Medium & \textbf{10.05} & \textbf{34.30} & \textbf{11.55} \\
        % TTAC++ (Contrastive Branch) \\
        \midrule
        BN~\cite{ioffe2015batch} & N-M-SF & Medium & 15.70 & 43.30 & 14.10 \\
        TENT~\cite{wang2020tent} & N-M-SF & Medium & 12.60 & 36.30 & 13.65 \\
        SHOT~\cite{pmlr-v119-liang20a} & N-M-SF & Medium & 14.70 & 38.10 & 14.25 \\
        TTAC++~(Ours) & N-M-SF & Medium & \textbf{9.14} & \textbf{34.43} & \textbf{10.60} \\
        \midrule
        TTT++~\cite{liu2021ttt++} & N-M-SL & Medium & 11.87 & 37.09 & 11.95 \\
        TTAC~\cite{su2022revisiting} & N-M-SL & Medium & 9.42 & 33.55 & 11.00 \\
        TTAC+SHOT~\cite{su2022revisiting} & N-M-SL & Medium & 9.54 & 32.89 & 11.30 \\
        TTAC++~(Ours) & N-M-SL & Medium & \textbf{7.23} & \textbf{29.23} & \textbf{9.00} \\
        \midrule
        TTT-R~\cite{sun2020test} & Y-M-SL & Strong & 14.30 & 40.40 & 11.00 \\ 
        TTT++~\cite{liu2021ttt++} & Y-M-SL & Strong & 9.80 & 34.10 & 9.50 \\
        TTAC~\cite{su2022revisiting} & Y-M-SL & Strong & 8.52 & 30.57 & 9.20 \\
        TTAC++~(Ours) & Y-M-SL & Strong & \textbf{7.57} & \textbf{29.08} & \textbf{8.90} \\
        \bottomrule
        \end{tabular}
    }
    \label{tab:categorization_table}
\end{table*}

\begin{table*}[ht]
    \caption{Test-time training on ImageNet-C under the sTTT~(N-O) protocol. }%TTAC$^\dagger$ indicates TTAC with weak augmentation.}
    \centering
    \resizebox{\linewidth}{!}{
        \begin{tabular}{c|ccccccccccccccc|c}
        \toprule
        Method & Brit & Contr & Defoc & Elast & Fog & Frost & Gauss & Glass & Impul & Jpeg & Motn & Pixel & Shot & Snow & Zoom & Avg \\
        \midrule
        TEST & 38.82 & 89.55 & 82.23 & 87.13 & 64.84 & 76.83 & 97.34 & 90.50 & 97.76 & 68.31 & 83.60 & 80.37 & 96.74 & 82.22 & 74.31 & 80.70 \\
        BN (N-O-SF) & 32.33 & 50.93 & 81.28 & 52.98 & 42.21 & 64.13 & 83.25 & 83.64 & 82.52 & 59.18 & 66.23 & 49.45 & 82.59 & 62.34 & 52.51 & 63.04 \\
        TENT (N-O-SF) & 31.39 & 40.27 & 75.68 & 42.03 & 35.38 & 64.32 & 84.92 & 84.96 & 81.43 & 46.84 & 49.48 & 39.77 & 84.21 & 49.23 & 43.49 & 56.89 \\
        SHOT (N-O-SF) & {30.69} & 37.69 & 61.97 & 41.30 & 34.74 & 54.19 & 76.33 & 71.94 & 74.24 & 46.50 & 47.98 & 38.88 & 70.60 & 46.09 & {40.74} & 51.59 \\
        Conjugate PL (N-O-SF) & \textbf{30.62} & \textbf{34.28} & {61.12} & {40.40} & {34.43} & {51.80} & {65.61} & {67.75} & {63.71} & \textbf{44.61} & {45.70} & \textbf{38.41} & {63.07} & {45.83} & 41.27 & {48.57} \\
        
        Self-Training (N-O-SF) & 31.57 & 37.62 & 79.68 & 42.84 & 35.27 & 54.18 & 88.76 & 91.93 & 81.22 & 51.97 & 50.29 & 39.73 & 88.67 & 48.52 & 47.07 & 57.95 \\
        TTAC++ (N-O-SF) & 31.61 & {36.55} & \textbf{60.39} & \textbf{38.79} & \textbf{34.20} & \textbf{49.02} & \textbf{61.62} & \textbf{62.67} & \textbf{59.37} & {45.25} & \textbf{44.73} & {38.43} & \textbf{58.32} & \textbf{43.60} & \textbf{39.99} & \textbf{46.97} \\
        \midrule
        
        TTAC (N-O-SL) & {30.36} & 38.84 & 69.06 & {39.67} & 36.01 & {50.20} & 66.18 & 70.17 & 64.36 & {45.59} & 51.77 & 39.72 & {62.43} & {44.56} & 42.80 & 50.11 \\
        
        %TTAC$^\dagger$ (N-O-SL) & 30.40 & \textbf{34.01} & {61.60} & 39.95 & {35.42} & 51.55 & {65.83} & {68.58} & {64.11} & 45.77 & {46.33} & {39.60} & 63.29 & 45.62 & {40.71} & {48.85} \\
        
        TTAC++ (N-O-SL) & \textbf{29.78} & \textbf{34.37} & \textbf{58.08} & \textbf{37.68} & \textbf{32.97} & \textbf{47.96} & \textbf{60.51} & \textbf{62.24} & \textbf{58.65} & \textbf{43.61} & \textbf{43.58} & \textbf{36.89} & \textbf{57.33} & \textbf{42.40} & \textbf{38.82} & \textbf{45.66} \\
        \bottomrule
    \end{tabular}
    }
    \label{tab:ImageNet}
    \vspace{-0.2cm}
\end{table*}

\subsubsection{Competing Methods}
We compare the following test-time training methods. Direct testing (\textbf{TEST}) without adaptation simply do inference on target domain with source domain model.
Test-time training (\textbf{TTT-R})~\cite{sun2020test} jointly trains the rotation-based self-supervised task and the classification task in the source domain, and then only train the rotation-based self-supervised task in the streaming test samples and make the predictions instantly. The default method is classified into the Y-M protocol.
Test-time normalization (\textbf{BN})~\cite{ioffe2015batch} moving average updates the batch normalization statistics by streamed data. The default method follows N-M protocol and can be adapted to N-O protocol.
Test-time entropy minimization (\textbf{TENT})~\cite{wang2020tent} updates the parameters of all batch normalization by minimizing the entropy of the model predictions in the streaming data. By default, TENT follows the N-O protocol and can be adapted to N-M protocol.
Test-time classifier adjustment (\textbf{T3A})~\cite{iwasawa2021test} computes target prototype representation for each category using streamed data and make predictions with updated prototypes. T3A follows the N-O protocol by default.
Source Hypothesis Transfer (\textbf{SHOT})~\cite{pmlr-v119-liang20a} freezes the linear classification head and trains the target-specific feature extraction module by exploiting balanced category assumption and self-supervised pseudo-labeling in the target domain. SHOT by default follows the N-M protocol and we adapt it to N-O protocol.
\textbf{TTT++}~\cite{liu2021ttt++} aligns source domain feature distribution, whose statistics are calculated offline, and target domain feature distribution by minimizing the F-norm between the mean covariance. TTT++ follows the Y-M protocol and we adapt it to N-O (removing contrastive learning branch) and Y-O protocols. 
\textbf{AdaContrast}~\cite{chen2022contrastive} approaches TTT from a self-training perspective. Pseudo labels on target domain testing samples are generated from a weak augmentation branch and used for supervising a strong augmentation branch.
\textbf{Conjugate PL}~\cite{goyaltest2022} proposed to learn the best TTT objective through meta-learning. This approach discovered a loss similar to the entropy loss adopted by Tent when source domain is trained with cross-entropy loss.
\textbf{Self-Training}~\cite{sohn2020fixmatch}, a.k.a. FixMatch, was originally developed for semi-supervised learning by estimating pseudo labels on the unlabeled data and supervise model training with pseudo labels. We adapt FixMatch to TTT by adopting the self-training component alone and refer to it as Self-Training~(ST).
\textbf{TTAC}~\cite{su2022revisiting} aligns the source and target domain feature distributions for TTT. It requires a single pass on the target domain and does not have to modify the source training objective. TTAC was originally implemented for all TTT protocols. TTAC was further augmented with additional diversity loss and entropy minimization loss introduced in SHOT~\cite{pmlr-v119-liang20a}, denoted as TTAC+SHOT~\cite{su2022revisiting}. Finally, we evaluate our proposed method, \textbf{TTAC++}, under all TTT protocols. For Y-O and Y-M protocols we incorporate an additional contrastive learning branch introduced in~\cite{liu2021ttt++}. %\textcolor{red}{For Yongyi: Please briefly introduce Conjugate PL and AdaContrast.}

% \vspace{-0.4cm}

\subsection{Test-Time Training Evaluations}\label{sect:ttt_eval}
We evaluate test-time training on four types of target domain data, including images with corruptions, manually selected hard images, synthetic to real adaptation and adversarial samples.

\subsubsection{TTT on Corrupted Target Domain}
We present the test-time training results on CIFAR10/100-C datasets in Tab.~\ref{tab:categorization_table}, {and the results on ImageNet-C dataset in Tab.~\ref{tab:ImageNet}.} For ImageNet-C, we only evaluate under the realistic sTTT~(N-O) protocol. We make the following observations from the results.

\noindent\textbf{sTTT (N-O) Protocol}. 
We first analyze the results under the proposed sTTT (N-O) protocol. Our method outperforms all competing ones by a large margin both with source domain statistics~(N-O-SL) and without source domain statistics~(N-O-SF). Under the most strict N-O-SF protocol, TTAC++ leads the benchmark with a large margin. It outperforms Conjugate PL by $1.6\%$ on CIFAR10-C and SHOT by $1.4\%$ on CIFAR100-C. When source domain statistics are available, TTAC++ gains additional advantage. Compared with TTT++, we achieved $4\%$ and $5\%$ improvements on CIFAR10-C and CIFAR100-C datasets respectively. TTAC++ also improves upon TTAC+SHOT where the latter adopts a class balance assumption. On ImageNet-C dataset, TTAC++ demonstrates its superiority under both N-O-SF and N-O-SL protocols. Without source domain statistics, TTAC++ outperforms Conjugate PL on 11 out of 15 types of corruptions. When source domain statitics are available, TTAC++ consistently outperforms TTAC with similar data augmentation.

% For example, $3\%$ improvement is observed on both CIFAR10-C and CIFAR100-C from the previous best (TTT++) under N-O-SL and
% {5-13\% improvement is observed on ImageNet-C compared with BN and TENT under N-O-SF, and TTAC++ is superior in average accuracy and outperforms on 11 out of 15 types of corruptions compared with ConjugatePL on ImageNet-C}. 

%We further combine TTAC with the class balance assumption made in SHOT (TTAC+SHOT). With the stronger assumptions out method can further improve upon TTAC alone, in particular on ModelNet40-C dataset. This result demonstrates TTAC's compatibility with existing methods.

\begin{table*}[htbp]
    \centering
    \caption{Test-time training on the VisDA-C dataset.} 
    % The numbers for competing methods are inherited from \cite{liu2021ttt++}.
    % \renewcommand{\arraystretch}{1.2}
          \setlength\tabcolsep{3pt} % default value: 6pt
    \resizebox{0.68\linewidth}{!}{
        \begin{tabular}{c|c|cccccccccccc|cc}
        \toprule
        Method & Protocol & Plane & Bcycl & Bus & Car & Horse & Knife & Mcycl & Person & Plant & Sktbrd & Train & Truck & Avg\\
        \midrule
        - & TEST & 56.52 & 88.71 & 62.77 & 30.56 & 81.88 & 99.03 & 17.53 & 95.85 & 51.66 & 77.86 & 20.44 & 99.51 & 65.19\\
        \midrule
        \multirow{4}{*}{N-O-SF} 
        & TENT &  19.75 & 81.99 & 17.78 & 40.03 & 21.64 & 19.04 & 11.66 & 38.18 & 23.15 & 77.33 & 35.88 & 98.31 & 40.40 \\
        & SHOT & 10.81 & \textbf{18.62} & 27.08 & 59.65 & 11.13 & 56.43 & 27.29 & 26.22 & 13.76 & 47.35 & 22.26 & \textbf{61.18} & 31.82 \\
        & Self-Training & \textbf{4.69} & 21.12 & \textbf{13.67} & \textbf{16.00} & \textbf{4.03} & 89.64 & \textbf{6.19} & 86.35 & \textbf{3.17} & 88.82 & \textbf{15.18} & 98.05 & 37.24 \\
        & TTAC++ & 11.63 & 22.42 & 18.49 & 33.47 & 9.57 & \textbf{18.89} & 10.28 & \textbf{21.92} & 10.75 & \textbf{24.46} & 18.51 & 90.37 & \textbf{24.23} \\
        \midrule
        
        \multirow{2}{*}{N-O-SL} 
        & TTAC & 18.54 & 40.20 & 35.84 & 63.11 & 23.83 & 39.61 & 15.51 & 41.35 & 22.97 & 46.56 & 25.24 & 67.81 & 36.71 \\
        %& TTAC$^\dagger$ & 22.87 & 41.15 & 34.14 & 61.78 & 24.13 & 50.46 & 16.86 & 42.08 & 25.02 & 45.51 & 23.04 & 67.77 & 37.90\\
        & TTAC++ & \textbf{7.13} & \textbf{31.34} & \textbf{21.79} & \textbf{43.07} & \textbf{7.57} & \textbf{13.25} & \textbf{7.52} & \textbf{27.95} & \textbf{8.33} & \textbf{32.00} & \textbf{14.16} & \textbf{63.66} & \textbf{23.15} \\
        \midrule
        
        \multirow{2}{*}{Y-O-SL} 
        & TTAC & 7.19 & 29.99 & 22.52 & 56.58 & 8.14 & 18.41 & 8.25 & 22.28 & 10.18 & \textbf{23.98} & 13.55 & \textbf{67.02} & 24.01 \\
        & TTAC++ & \textbf{4.85} & \textbf{26.45} & \textbf{20.98} & \textbf{44.01} & \textbf{5.41} & \textbf{7.47} & \textbf{6.90} & \textbf{21.95} & \textbf{6.53} & 27.49 & \textbf{12.58} & 68.91 & \textbf{21.13}\\
        \midrule
        
        \multirow{6}{*}{N-M-SF} 
        & BN & 44.38 & 56.98 & 33.24 & 55.28 & 37.45 & 66.60 & 16.55 & 59.02 & 43.55 & 60.72 & 31.07 & 82.98 & 48.99\\
        & TENT & 13.43 & 77.98 & 20.17 & 48.15 & 21.72 & 82.45 & 12.37 & 35.78 & 21.06 & 76.41 & 34.11 & 98.93 & 45.21\\
        & SHOT & 5.73 & \textbf{13.64} & 23.33 & 42.69 & 7.93 & 86.99 & 19.17 & 19.97 & 11.63 & \textbf{11.09} & 15.06 & \textbf{43.26} & 25.04 \\
        & Self-Training & 4.44 & 26.91 & 16.25 & \textbf{22.87} & \textbf{3.45} & 60.53 & \textbf{5.59} & 53.50 & \textbf{4.31} & 61.25 & 17.68 & 95.04 & 30.99 \\
        & AdaContrast & \textbf{4.28} & 14.65 & 22.52 & 27.58 & 4.24 & \textbf{7.23} & 13.77 & \textbf{13.00} & 6.05 & 87.46 & \textbf{9.28} & 51.98 & 21.84 \\
        & TTAC++ & 7.62 & 17.93 & \textbf{15.69} & 27.17 & 5.27 & 9.73 & 7.32 & 21.28 & 7.19 & 21.35 & 15.18 & 92.93 & \textbf{20.72} \\
        \midrule

        \multirow{3}{*}{N-M-SL} 
        & TTT++ & 28.25 & 32.03 & 33.67 & 64.77 & 20.49 & 56.63 & 22.52 & 36.30 & 24.84 & 35.20 & 25.31 & 64.24 & 37.02 \\
        & TTAC & 14.43 & 36.52 & 34.90 & 61.94 & 21.34 & 45.06 & 13.41 & 39.12 & 17.48 & 42.83 & 25.24 & 65.36 & 34.80 \\
        %& TTAC$^\dagger$ & 18.02 & 34.27 & 30.47 & 58.74 & 19.42 & 54.41 & 14.82 & 39.58 & 16.42 & 41.60 & 25.14 & 65.50 & 34.87\\
        & TTAC++ & \textbf{5.46} & \textbf{27.02} & \textbf{18.14} & \textbf{38.19} & \textbf{5.69} & \textbf{11.47} & \textbf{7.18} & \textbf{28.77} & \textbf{7.50} & \textbf{13.28} & \textbf{13.17} & \textbf{59.82} & \textbf{19.64} \\
        \midrule
        
        \multirow{3}{*}{Y-M-SL} 
        & TTT++ & 4.13 & 26.20 & 21.60 & \textbf{31.70} & 7.43 & 83.30 & 7.83 & 21.10 & 7.03 & 7.73 & \textbf{6.91} & \textbf{51.40} & 23.03\\
        & TTAC & 2.74 & 17.73 & 18.91 & 43.12 & 5.54 & 12.24 & \textbf{4.66} & \textbf{15.90} & 4.77 & 10.78 & 9.75 & 62.45 & 17.38 \\
        & TTAC++ & \textbf{2.61} & \textbf{16.86} & \textbf{16.82} & 38.41 & \textbf{4.28} & \textbf{2.89} & 4.93 & 18.20 & \textbf{4.29} & \textbf{6.27} & 8.78 & 62.76 & \textbf{15.59}\\
        \bottomrule
        \end{tabular}
    }
    \label{tab:visda}
\end{table*}

\noindent\textbf{Alternative Protocols}.
We further compare different methods under N-M, Y-O and Y-M protocols. Under the Y-O protocol, TTT++~\cite{liu2021ttt++} modifies the source domain training objective by incorporating a contrastive learning branch~\cite{chen2020simple}. To compare with TTT++, we also include the contrastive branch and observe a clear improvement on both CIFAR10-C and CIFAR100-C datasets. Other TTT methods are adapted to the N-M protocol by allowing training on the whole target domain data multiple epochs. Specifically, we compared with BN, TENT and SHOT. With TTAC alone we observe substantial improvement on all three datasets and TTAC can be further combined with SHOT demonstrating additional improvement. Finally, under the Y-M protocol, we demonstrate very strong performance compared to TTT-R and TTT++. It is also worth noting that TTAC under the N-O protocol can already yield results close to TTT++ under the Y-M protocol, suggesting the strong test-time training ability of TTAC even under the most challenging TTT protocol.

\subsubsection{TTT on Selected Hard Samples as Target Domain}

CIFAR10.1 contains roughly 2,000 new test images that were re-sampled after the research on original CIFAR-10 dataset, which consists of some hard samples and reflects the normal domain shift in our life. Evaluation of TTT methods on CIFAR10.1 is widely adopted to verify the benefit of adapting to hard target domain samples. We present the results on CIFAR10.1 in Table.~\ref{tab:categorization_table}. Again, we observe a strong performance of TTAC++ under all TTT protocols. 

% Describe CIFAR-10.1

\subsubsection{TTT on Synthetic Source to Real Target Domains}

VisDA-C is a large-scale benchmark of synthetic-to-real object classification dataset. We demonstrate on this dataset the ability of TTT to adapt model trained on synthetic source domain to realistic target domain data. %The setting of training on a synthetic dataset and testing on real data fits well with the real application scenario. 
On this dataset, we conduct experiments under the N-O, Y-O, N-M and Y-M protocols with results presented in Table.~\ref{tab:visda}. We make the following observations from the results. First, our proposed method, TTAC++, outperforms all competing methods under all evaluation protocols. In particular, the improvement for TTAC++ is very significant under the N-O~(sTTT) protocol regardless of access to source domain statistics. For example, TTAC++ outperforms the previous best method, the SHOT, by $7\%$ under N-O-SF and, the TTAC, by $13\%$ under N-O-SL. Moreover, TTAC++ also demonstrates very competitive performance when multiple passes on the target domain is allowed~(N-M-SF), a.k.a. source-free domain adaptation.

\subsubsection{TTT on Adversarial Target Domain}

The existing test-time training/adaptation works often evaluate adaptation to the target domain with hand-crafted corruptions. In this section, we investigate the robustness of test-time training subject to stronger out-of-distribution data, i.e. adversarial samples.

This evaluation reveals that simple test-time training can substantially improve model's robustness to adversarial testing samples. In specific, we generate adversarial samples on the testing set of CIFAR-10 dataset by $L^\infty$ PGD~\cite{madry2018towards} attack with $\epsilon=8/255$, 40 iterations and attack step size $0.01$. The adversarial testing samples are then frozen for TTT evaluation. We extensively evaluated existing TTT methods and present the results in Tab.~\ref{tab:AdversarialSample}. We make the following observations from the results. First, without any test-time adaptation, direct testing with source domain model yields very poor performance ($93.07\%$). This is consistent with previous investigations into adversarial attacks. Furthermore, existing test-time training methods which do not consider self-training, e.g. BN, TENT and SHOT, performs relatively poor compared with methods equipped with self-training, e.g. Self-Training, TTT++ and TTAC++. We attribute the performance gap to the fact that the data augmentation applied during self-training is able to smooth out the adversarial noise. Self-training is able to exploit the pseudo labels predicted on smoothed testing samples and improve the accuracy. Finally, we observe that distribution matching is complementary to purely self-training, suggested by the improvement of TTAC++ from Self-Training under N-O-SF and N-M-SF protocols. Overall, we demonstrate that TTAC++ is a strong test-time training paradigm, it owns the ability to adapt to adversarial corruption which is stronger than hand-crafted natural corruptions.

% \noindent\textbf{TTT on Adversarial Samples}. \textcolor{red}{For Yongyi: Please describe how we generated the adversarial samples.}
% To evaluate the robustness on adversarial samples, we generate the adversarial samples on CIFAR10 test set by Projected Gradient Descent (PGD)~\cite{madry2018towards}. Specifically, referring to Advertorch\footnote{https://github.com/BorealisAI/advertorch}, we leverage the $L^\infty$ order PGD attack algorithm and set the maximum distortion $\epsilon$ to $8/255$, the number of iterations to $40$, and the attack step size to $0.01$.

\begin{table*}[htbp]
    \centering
    \caption{Evaluations on test-time training for adversarial samples of CIFAR10 dataset. Each number in the table indicates error rate in percentage.} %TTAC$^\dagger$ aligns source and target distribution with weakly augmented samples.}
          \setlength\tabcolsep{3pt} % default value: 6pt
    \resizebox{0.61\linewidth}{!}{
    
        \begin{tabular}{c|c|cccccccccc|c}
            \toprule
            Protocol & Method & Airpl. & Automob. & Bird & Cat & Deer & Dog & Frog & Horse & Ship & Truck & Avg \\
            \midrule
            - & TEST & 97.00 & 84.90 & 95.20 & 93.20 & 95.70 & 96.30 & 95.00 & 95.70 & 88.80 & 88.90 & 93.07 \\
            \midrule
            \multirow{5}{*}{N-O-SF} 
            & BN & 95.80 & 62.30 & 89.50 & 90.60 & 91.70 & 79.60 & 78.10 & 73.60 & 83.10 & 67.20 & 81.15 \\
            & TENT & 96.70 & 65.60 & 92.10 & 92.80 & 93.20 & 86.40 & 82.40 & 78.70 & 90.50 & 73.00 & 85.14\\
            & SHOT & 97.50 & 50.60 & 90.80 & 93.50 & 93.90 & 74.30 & 78.00 & 67.60 & 85.00 & 65.50 & 79.67\\
            & Self-Training & {43.10} & \textbf{12.20} & {48.20} & \textbf{53.00} & {45.30} & {45.90} & 48.40 & 42.30 & {29.20} & {35.00} & {40.26} \\
            & TTAC++ & \textbf{36.70} & 14.70 & \textbf{45.80} & 60.70 & \textbf{37.80} & \textbf{40.70} & \textbf{23.40} & \textbf{33.30} & \textbf{25.60} & \textbf{27.60} & \textbf{34.63} \\
            \midrule
            \multirow{3}{*}{N-O-SL} 
            & TTT++ & 89.60 & 46.60 & 79.90 & 86.00 & 83.80 & 66.40 & 60.80 & 47.90 & 72.70 & 56.10 & 68.98\\
            & TTAC & 60.20 & 26.30 & 58.60 & 68.20 & 62.10 & 47.10 & {41.40} & {29.70} & 36.90 & 37.20 & 46.77\\
            %TTAC$^\dagger$ & N-O-SL & 54.60 & 32.90 & 60.40 & 74.20 & 60.80 & 62.90 & 44.20 & 42.40 & 39.10 & 39.90 & 51.14 \\
            & TTAC++ & \textbf{23.90} & \textbf{12.60} & \textbf{32.00} & \textbf{53.50} & \textbf{34.60} & \textbf{33.90} & \textbf{20.20} & \textbf{15.90} & \textbf{17.20} & \textbf{18.80} & \textbf{26.26} \\
            \midrule
            \multirow{3}{*}{Y-O-SL} 
            & TTT++ & 37.80 & 15.00 & 47.70 & 58.50 & 41.80 & 38.60 & 26.10 & 20.20 & 19.00 & 21.90 & 32.66 \\
            & TTAC & 32.40 & 14.50 & 42.60 & 58.80 & 42.10 & 37.50 & 23.20 & 18.70 & 16.00 & 23.00 & 30.88\\
            %TTAC$^\dagger$ & Y-O-SL & 31.00 & 17.10 & 39.80 & 56.80 & 41.20 & 40.80 & 26.30 & 24.00 & 15.70 & 21.60 & 31.43\\
            & TTAC++ & \textbf{23.30} & \textbf{11.20} & \textbf{32.30} & \textbf{54.10} & \textbf{31.70} & \textbf{30.90} & \textbf{17.80} & \textbf{15.10} & \textbf{14.70} & \textbf{18.60} & \textbf{24.97}\\
            \midrule
            \multirow{5}{*}{N-M-SF} 
            & BN & 96.20 & 59.10 & 87.90 & 90.90 & 90.70 & 79.80 & 74.80 & 72.30 & 82.90 & 64.70 & 79.93 \\
            & TENT & 96.00 & 66.60 & 91.50 & 92.20 & 93.90 & 83.20 & 81.20 &	76.90 &	87.10 &	71.50 &	84.01 \\
            & SHOT & 96.70 & 48.10 & 89.50 & 93.50 & 93.00 & 75.30 & 78.60 & 43.40 & 80.60 & 64.70 & 76.34 \\
            & Self-Training & 30.70 & \textbf{4.40} & 35.90 & 54.30 & 39.10 & \textbf{18.50} & 45.80 & \textbf{30.30} & \textbf{19.90} & 26.90 & 30.58 \\
            & TTAC++ & \textbf{23.10} & {8.10} & \textbf{36.60} & \textbf{36.90} & \textbf{29.50} & 41.20 & \textbf{16.90} & 50.30 & 26.50 & \textbf{20.30} & \textbf{28.94}\\
            \midrule
            \multirow{3}{*}{N-M-SL} 
            & TTT++ & 74.90 & 28.50 & 68.00 & 78.30 & 71.40 & 51.70 & 47.70 & 31.10 & 52.70 & 40.90 & 54.52 \\
            & TTAC & 36.60 & 15.10 & 46.30 & 54.40 & 47.60 & 39.10 & 28.90 & 21.90 & 21.30 & 22.10 & 33.33 \\
            %TTAC$^\dagger$ & N-M-SL & {14.80} & 8.30 & {23.90} & {35.50} & {23.90} & {26.40} & {14.50} & {13.30} & {9.10} & {13.30} & {18.30} \\
            & TTAC++ & \textbf{11.10} & \textbf{5.60} & \textbf{15.40} & \textbf{28.90} & \textbf{14.50} & \textbf{17.40} & \textbf{9.30} & \textbf{7.00} &	\textbf{6.20} & \textbf{10.10} & \textbf{12.55} \\
            \midrule
            \multirow{3}{*}{Y-M-SL} 
            & TTT++ & 19.30 & 6.80 & 30.10 & 41.90 & 25.60 & 27.50 & 14.20 & 10.60 & 7.80 & 11.60 & 19.54\\
            % TTT++ & Y-M-SL & \\
            & TTAC & 18.50 & 6.90 & 31.40 & 43.20 & 29.00 & 32.00 & 16.50 & 11.70 & 6.90 & 12.20 & 20.83\\
            %TTAC$^\dagger$ & Y-M-SL & 11.90 & 5.90 & 19.70 & 30.40 & 17.60 & 21.00 & 10.60 & 9.80 & 6.70 & 10.30 & 14.39\\
            & TTAC++ & \textbf{11.20} & \textbf{3.90} & \textbf{15.70} & \textbf{26.50} & \textbf{14.50} & \textbf{17.10} & \textbf{7.60} & \textbf{7.40} & \textbf{5.60} & \textbf{9.50} & \textbf{11.90}\\
            \bottomrule
        \end{tabular}
    }
    
    \label{tab:AdversarialSample}
\end{table*}

% \begin{table}[htbp]
%     \centering
%     \caption{Caption}
%     \begin{tabular}{c|ccccccc}
%         \toprule
%         Protocol & TEST & BN & TENT & SHOT & TTT++ & TTAC & TTAC++ \\
%         \midrule
%         N-O & \\
%         N-M & \\
%         \bottomrule
%     \end{tabular}
%     \label{tab:AdversarialSample}
% \end{table}

\subsubsection{TTT Cumulative Performance}
\begin{figure*}[!htb]
    \centering
    \includegraphics[width=0.9\linewidth]{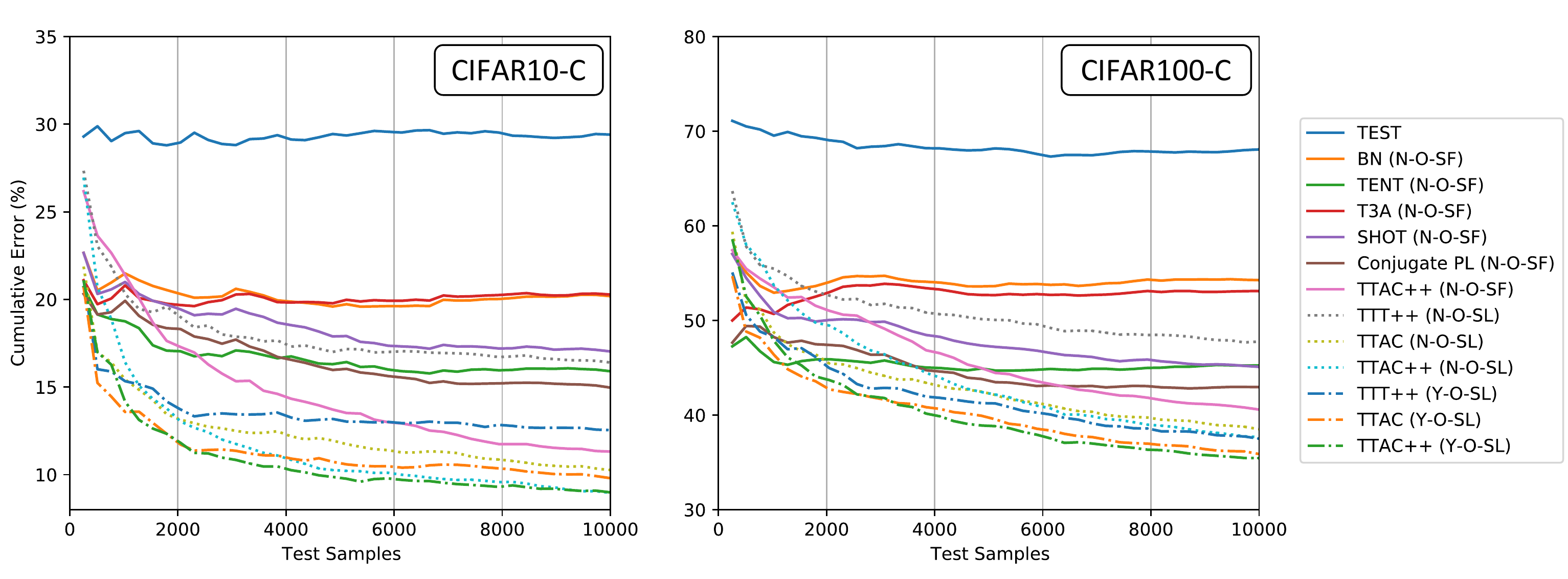}
    \caption{Test-time cumulative error rate on CIFAR10/100-C datasets.}
    \label{fig:cifar_cumulative}
\end{figure*}

\begin{figure}
    \centering
    \includegraphics[width=\linewidth]{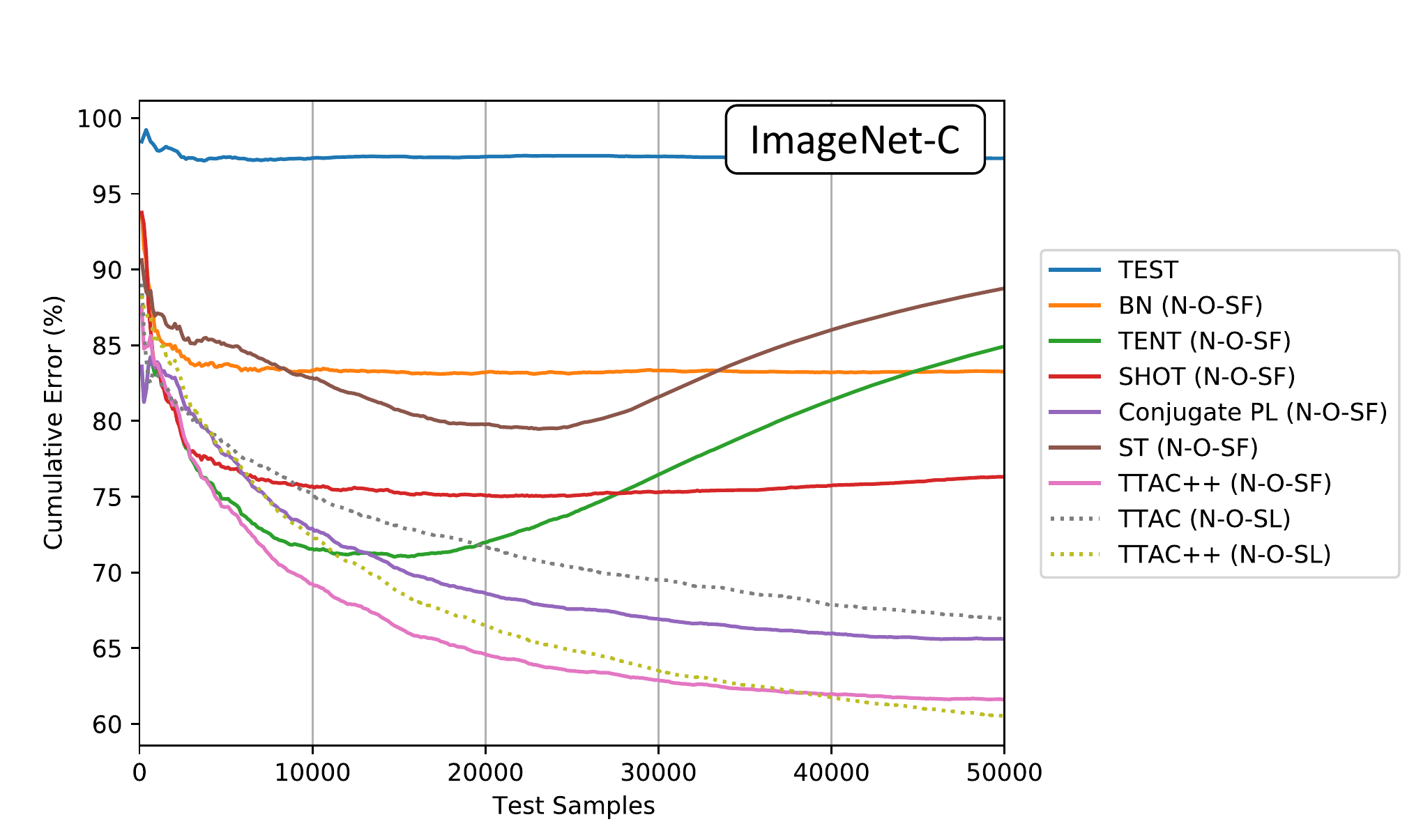}
    \caption{Test-time cumulative error on ImageNet-C dataset with Gaussian Blur corruption.}
    \label{fig:imagenet_cumulative}
\end{figure}

A good test-time training framework should benefit from  seeing more testing samples and the performance on the target domain is expected to be gradually increasing. In this section, we compare different test-time training methods by illustrating the cumulative error rate on CIFAR10/100-C and ImageNet-C~(\textit{Gaussian Blur} corruption) under the sTTT protocols~(N-O-SF/SL) in Fig.~\ref{fig:cifar_cumulative} and Fig.~\ref{fig:imagenet_cumulative}, respectively. As we observe from the figure, some existing TTT methods do not benefit from seeing more testing samples.  For example, BN and T3A's performance stabilize after 2000 testing samples on CIFAR10/100-C.  The performance of TENT and Self-Training~(ST) even degrade after observing 10000 to 20000 testing samples on ImageNet-C. This empirical evaluation also suggest applying self-training alone for TTT is prone to confirmation bias and may harm the performance. In contrast, TTAC++~(pink solid line) exhibits the fastest drop of error rate among all source-free methods. Compared with all methods requiring access to source domain statistics or changing source domain training objectives, TTAC++ is is consistently lower in error rate along the TTT procedure. More importantly, the trend~(sharp slope) shows higher potential for TTAC++ should more testing samples are available in the target domain.

 % For both datasets TTAC outperforms competing methods from the early stage of test-time training. The advantage is consistent throughout the TTT procedure.

\begin{figure}
    \subfloat[TTAC Feature]{\includegraphics[width=0.5\linewidth]{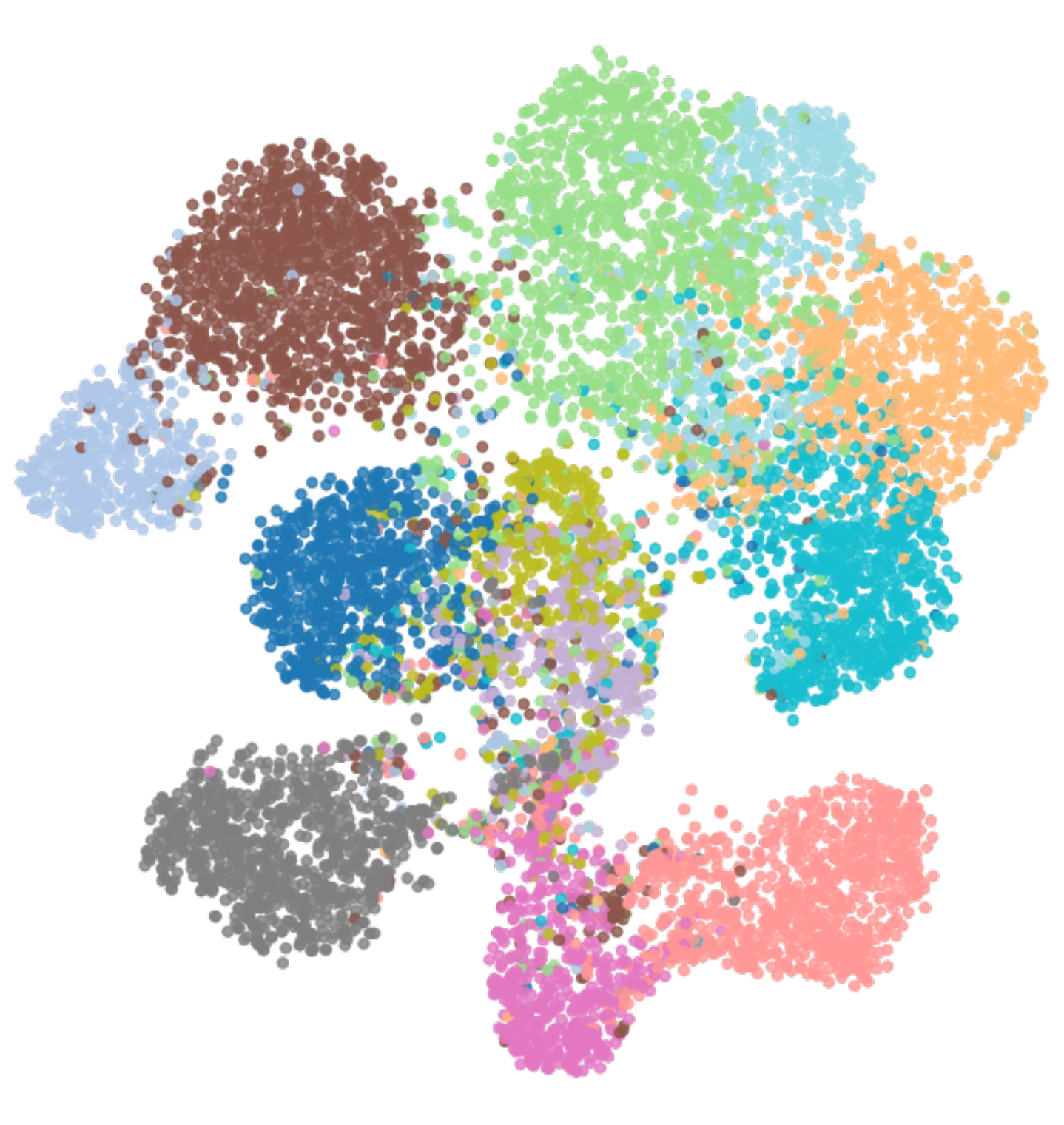}}
    \subfloat[TTAC++ Feature]{\includegraphics[width=0.5\linewidth]{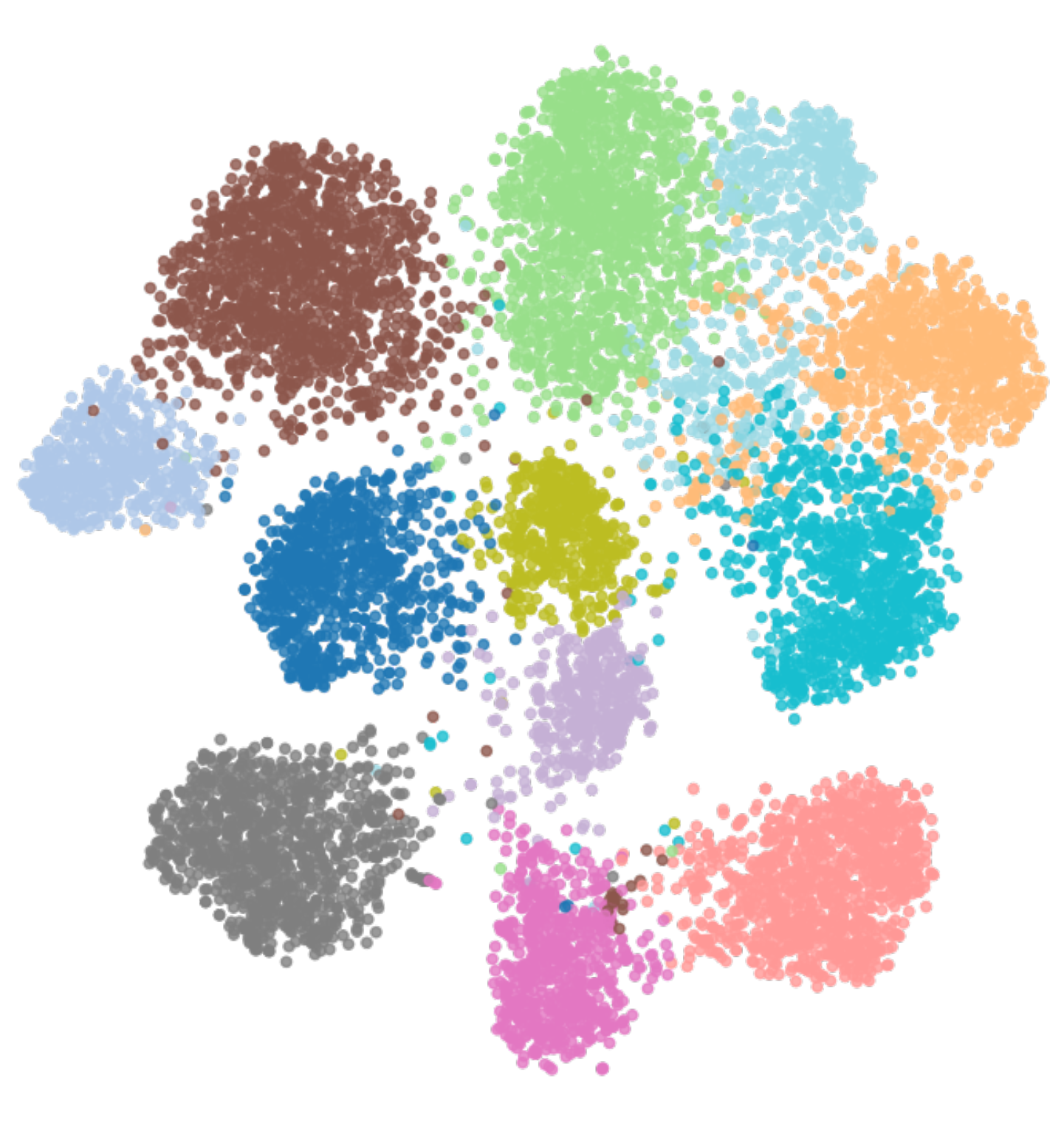}}

    \centering
    \caption{To reduce the computation, we select 10,000 samples on VisDA-C dataset to draw the T-SNE visualizations. (a) T-SNE visualization of TTAC feature embedding. (b) T-SNE visualization of TTAC++ feature embedding.}\label{fig:tsne}
\end{figure}

\subsubsection{TSNE Visualization of TTAC++ features}
We provide qualitative results for test-time training by visualizing the adapted features through T-SNE~\cite{van2008visualizing}. In Fig.~\ref{fig:tsne}~(a) and Fig.~\ref{fig:tsne}~(b), we compare the features learned by TTAC~\cite{su2022revisiting} and TTAC++. We observe a better separation between classes by TTAC++, implying an improved classification accuracy.

% \newpage
% \vspace{-0.4cm}
\subsection{Ablation Study}
% \vspace{-0.1cm}
% We ablate the models under two TTT settings ...
% Introduce A.C., PLF and GA. Analyze table 4

In this section, we validate the effectiveness of individual components, including anchored clustering, pseudo label filtering, global feature alignment, self-training and finally the compatibility with contrastive branch~\cite{liu2021ttt++}, on CIFAR10-C dataset.
%We conduct ablation study on CIFAR10-C dataset for individual components, including anchored clustering, pseudo label filtering, global feature alignment and finally the compatibility with contrastive branch. 
For anchored clustering alone, we use all testing samples to update cluster statistics. For pseudo label filtering alone, we implement as predicting pseudo labels followed by filtering, then pseudo labels are used for self-training. We make the following observations from Tab.~\ref{tab:ablation}. Under both N-O and N-M protocols, introducing anchored clustering or pseudo label filtering alone improves over the baseline, e.g. under N-O $29.15\%\rightarrow 14.32\%$ for anchored clustering and $29.15\%\rightarrow15.00\%$ for pseudo label filtering. When anchored clustering is combined with pseudo label filtering, we observe a significant boost in performance. This is due to more accurate estimation of category-wise cluster in the target domain.  We further evaluate aligning global features alone with KL-Divergence. This achieves relatively good performance and obviously outperforms the L2 distance alignment adopted in \cite{liu2021ttt++}. Next, when self-training is turned on in conjunction with other components, we observe a consistent improvement under all TTT protocols. Finally, we combine all components with additional contrastive branch and the full model yields the best performance under both Y-O-SL and Y-M-SL protocols. %When contrast learning branch is included, TTAC achieves even better results.

%To analyse the effectiveness of each module, including Anchor Clustering (A.C.), Pseudo Label Filter (P.L.F.) and Global Alignment (G.A.), we conduct adequate ablation experiments under the N-O and N-M setting on CIFAR10-C. In Table.~\ref{tab:ablation_study}, we firstly compare the results of the different module combinations under the N-O setting, where EM Algo. in the column of P.L.F. indicates that we utilize an EM algorithm to update the GMMs of target domain feature, only pseudo label filter indicates directly using pseudo label after filter to perform self-training. Under the N-M setting, we compare the use of KL-Divergence in global feature alignment module with the L2 distance leveraged in TTT++, and we can observe that the KL-Divergence is superior for handling distribution alignment tasks. Additionally, we observe that the anchored clustering module can't be performed without pseudo label filter and all the modules together can have a mutually reinforcing effect and get the optimal result.

% KLDivergence
% Sym KLD
% Multi Gaussian
% Pseudo Labeling

% Below table shows the differences between TTT++'s feature alignment module and KL Divergence module, between Single Gaussian Distribution and Multi Gaussian Distribution.
% TEST
% 

% \vspace{-0.5cm}

% Table generated by Excel2LaTeX from sheet 'Sheet2'
\begin{table*}[htbp]
  \centering
  \caption{Ablation study for individual components on CIFAR10-C dataset.}
    \resizebox{1\linewidth}{!}{
    \begin{tabular}{c|c|cccccc|cc|cccccc|cc}
    \toprule
    TTT Protocol & 
         \multicolumn{1}{c}{}
    & \multicolumn{6}{c}{N-O-SL} 
    & \multicolumn{2}{c}{Y-O-SL} 
    & \multicolumn{6}{c}{N-M-SL} 
    & \multicolumn{2}{c}{Y-M-SL}\\
    
    % \cmidrule(lr){1-1} \cmidrule(lr){2-2} \cmidrule(lr){3-8} \cmidrule(lr){9-10} \cmidrule(lr){11-16} \cmidrule(lr){17-18}
    \midrule
    
    Anchored Cluster. &
    -     
    & \checkmark & -     & \checkmark & -     & \checkmark & \checkmark
    & \checkmark & \checkmark
    & \checkmark & \checkmark & - & - & \checkmark & \checkmark
    & \checkmark & \checkmark\\
    
    Pseudo Label Filter. & 
    -     
    & - & \checkmark & \checkmark & - & \checkmark & \checkmark
    & \checkmark & \checkmark
    & - & \checkmark & - & - & \checkmark & \checkmark
    & \checkmark & \checkmark\\
    
    Global Feat. Align. & 
    - 
    & - & - & - & KLD & KLD & KLD
    & KLD & KLD
    & - & - & L2 Dist.\cite{liu2021ttt++} & KLD & KLD & KLD
    & KLD & KLD \\
    
    Self-Training & 
    -
    & - & - & - & - & - & \checkmark
    & - & \checkmark
    & - & - & - & - & - & \checkmark
    & - & \checkmark\\
    
    Contrast. Branch~\cite{liu2021ttt++} & 
    -
    & - & - & - & - & - & -
    & \checkmark & \checkmark
    & - & - & - & - & - & -
    & \checkmark & \checkmark \\
    
    Avg Acc & 
    29.15 
    & 14.32 & 15.00 & 11.33 & 11.72 & 10.94 & \textbf{9.78}
    & 10.69 & \textbf{10.05}
    & 11.11 & 10.01 & 11.87 & 10.8  & 9.42 & \textbf{7.23}
    & 8.52 & \textbf{7.57} \\
    \bottomrule
    \end{tabular}%
    }
  \label{tab:ablation}%
\end{table*}%

\subsection{Additional Analysis}

In this section, we provide additional investigations into additional the designs that affect computation efficiency, compatibility with additional backbones, randomness, and alternative designs, etc.

\begin{table*}[!htb]
    \centering
    \caption{Evaluating compatibility with ViT backbone on CIFAR10-C dataset.}
    \resizebox{\linewidth}{!}{
        \begin{tabular}{l|ccccccccccccccc|cc}
        \toprule
            Method & Brit & Contr & Defoc & Elast & Fog & Frost & Gauss & Glass & Impul & Jpeg & Motn & Pixel & Shot & Snow & Zoom & Avg & Std\\
        \midrule
            TEST & 2.29 & 16.24 & 4.83 & 9.45 & 13.60 & 6.73 & 24.52 & 18.23 & 24.48 & 12.63 & 7.63 & 14.57 & 23.02 & 5.29 & 3.50 & 12.47 & 7.36 \\
            BN    & 2.29 & 16.24 & 4.83 & 9.45 & 13.60 & 6.73 & 24.52 & 18.23 & 24.48 & 12.63 & 7.63 & 14.57 & 23.02 & 5.29 & 3.50 & 12.47 & 7.36  \\
            TENT  & 1.84 & 3.55  & \textbf{3.31} & 7.01 & 5.57 & 4.09 & 60.97 & 10.20 & 61.12 & 9.72  & 4.93 & 3.87  & 22.47 & 4.55 & 2.64 & 13.72 & 19.19 \\
            SHOT  & 2.00 & 3.13  & 3.46 & 6.63 & 5.79  & 4.06 & 11.65 & 9.39  & 10.58 & 9.69  & 5.03 & 3.63  & 10.05 & 4.35 & 2.70 & 6.14  & 3.15  \\
            TTT++ & 1.91 & 4.14  & 3.88 & 6.58 & 6.27  & 4.00    & 10.08 & 8.59  & 8.85  & 9.66  & 4.68 & \textbf{3.62} & 9.17  & 4.28 & 2.74 & 5.90  & 2.64  \\
            TTAC & 2.15 & 4.05  & 3.91 & 6.62 & 5.67  & 3.75 & 9.26  & 7.95  & \textbf{7.97}  & 8.55 & 4.75 & 3.87  & 8.24  & 3.93 & 2.94 & 5.57 & \textbf{2.24} \\
            TTAC++ & \textbf{1.59} & \textbf{3.01} & 3.70 & \textbf{6.19} & \textbf{4.45} & \textbf{3.25} & \textbf{9.14} & \textbf{7.76} & 8.02 & \textbf{7.84} & \textbf{4.14} & 3.64 & \textbf{7.80} & \textbf{3.38} & \textbf{2.47} & \textbf{5.09} & 2.36\\
        \bottomrule
        \end{tabular}
    }
    \label{tab:ViT}
\end{table*}

\subsubsection{Test Sample Queue and Update Epochs.}
{
Under the sTTT protocol, we allow all competing methods to maintain the same test sample queue and multiple update epochs on the queue. To analyze the significance of the sample queue and update epochs, we evaluate BN, TENT, SHOT, TTAC and TTAC++ on CIFAR10-C and ImageNet-C level 5 snow corruption evaluation set under different number of update epochs on test sample queue and under a without queue protocol, i.e. only update model w.r.t. the current test sample batch. As the results presented in Tab.~\ref{tab:QueueAccuracyAnalysis}, we make the following observations. i) Maintaining a sample queue can substantially improve the performance of methods that estimate target distribution, e.g. TTAC++~($11.18\rightarrow10.31$), TTAC~($11.91\rightarrow10.88$ on CIFAR10-C) and SHOT~($15.18\rightarrow13.96$ on CIFAR10-C). This is due to more test samples giving a better estimation of true distribution. ii) Consistent improvement can be observed with increasing update epochs for SHOT, TTAC and TTAC++. We ascribe this to iterative pseudo labeling benefiting from more update epochs. These observations also provide insights for deploying TTT for real-world practice. By considering memory constraint and demand for real-time model update, one can adjust the queue length and number of update epochs to strike a balance between efficiency and performance.
}

\begin{table}[!ht]
% \vspace{-0.4cm}
    \caption{Comparing with and without test sample queue and different numbers of model update epochs. w/ Queue maintains a test sample queue with 4096 samples; w/o Queue maintains a single mini-batch with 256 and 128 samples on CIFAR10-C and ImageNet-C respectively.}
    \centering
    \resizebox{0.99\linewidth}{!}{
    \begin{tabular}{c|cccc|c|cc|c}
    \toprule
         & \multicolumn{5}{c|}{CIFAR10-C} & \multicolumn{3}{c}{ImageNet-C} \\
    \midrule
         & \multicolumn{4}{c|}{w/ Queue } & \multicolumn{1}{c|}{w/o Queue} & \multicolumn{2}{c|}{w/ Queue} & w/o Queue \\
    \midrule
        \#Epochs & 1 & 2 & 3 & 4* & 1 & 1 & 2* & 1 \\
    \midrule
        BN & 15.84 & 15.99 & 16.04 & 16.00 & 15.44 & 62.34 & 62.34 & 62.59\\
        TENT & 13.35 & 13.83 & 13.85 & 13.87 & 13.48 & 47.82 & 49.23 & 48.39\\
        SHOT & 13.96 & 13.93 & 13.83 & 13.75 & 15.18 & 46.91 & 46.09 & 51.46\\
        TTAC & 10.88 & 10.80 & 10.58 & 10.01 & 11.91 & 45.44 & 44.56 & 46.64\\
        TTAC++ & \textbf{10.31} & \textbf{9.40} & \textbf{9.08} & \textbf{8.82} & \textbf{11.18} & \textbf{43.49} & \textbf{42.40} & \textbf{46.01}\\
    \bottomrule
    \end{tabular}
    }
    \label{tab:QueueAccuracyAnalysis}
\end{table}

\begin{table}[!ht]
% \vspace{-0.3cm}
    \caption{The per-sample wall time (measured in seconds) on CIFAR10-C under sTTT protocol.}
    \centering
    \resizebox{0.95\linewidth}{!}{
    \begin{tabular}{c|cccc|c}
    \toprule
         & \multicolumn{4}{c|}{w/ Queue } & \multicolumn{1}{c}{w/o Queue}\\
    \midrule
        \#Epochs & 1 & 2 & 3 & 4 & 1 \\
    \midrule
        BN & 0.0136 & 0.0220 & 0.0293 & 0.0362 & 0.0030\\
        TENT & 0.0269 & 0.0399 & 0.0537 & 0.0663 & 0.0041\\
        SHOT & 0.0479 & 0.0709 & 0.0942 & 0.1183 & 0.0067\\
        TTAC & 0.0516 & 0.0822 & 0.1233 & 0.1524 & 0.0083\\
        TTAC++ & 0.0706 & 0.1076 & 0.1591 & 0.1963 & 0.0090 \\
    \midrule
        Inference & 0.0030 & 0.0030 & 0.0030 & 0.0030 & 0.0030\\
    \bottomrule
    \end{tabular}
    }
    \label{tab:QueueTimeAnalysis}
\end{table}

\begin{table*}[!htb]
    \centering
    \caption{The performance of TTAC under different data streaming orders. }
    \resizebox{0.75\linewidth}{!}{
        \begin{tabular}{l|cccccccccc|cc}
        \toprule
           Random Seed & 0 & 10 & 20 & 200 & 300 & 3000 & 4000 & 40000 & 50000 & 500000 & Avg \\
        \midrule
           Error ($\%$) & 8.82 & 8.80 & 9.38 & 9.13 & 8.88 & 8.87 & 9.07 & 8.93 & 9.02 & 8.68 & 8.96$\pm$0.19 \\
        \bottomrule
        \end{tabular}
    }
    \label{tab:stream_order}
\end{table*}

\begin{table*}[!htb]
    \centering
    \caption{Comparison of alternative strategies for updating target domain clusters.  }
    \resizebox{0.9\linewidth}{!}{
        \begin{tabular}{l|ccccccccccccccc|c}
        \toprule
            Strategy & Brit & Contr & Defoc & Elast & Fog & Frost & Gauss & Glass & Impul & Jpeg & Motn & Pixel & Shot & Snow & Zoom & Avg \\
        \midrule
            i. Without filtering & 6.01 & 7.21 & 8.13 & 13.87 & 9.03 & 9.82 & 13.13 & 18.22 & 15.66 & 11.47 & 9.26 & 9.29 & 11.68 & 9.19 & 6.79 & 10.58 \\
            ii. Soft Assignment & 5.91 & 6.52 & 8.05 & 13.25 & 9.08 & 9.76 & 13.14 & 17.19 & 15.45 & 11.41 & 8.88 & 9.10 & 11.53 & 9.13 & 6.83 & 10.35 \\
            Filtering~(Ours)  & \textbf{5.59} & \textbf{6.28} & \textbf{7.53} & \textbf{12.99} & \textbf{8.95} & \textbf{9.22} & \textbf{12.13} & \textbf{15.79} & \textbf{14.37} & \textbf{10.65} & \textbf{8.70} & \textbf{8.60} & \textbf{10.70} & \textbf{8.82} & \textbf{6.37} & \textbf{9.78}\\
        \bottomrule
        \end{tabular}
    }
    \label{tab:target_clust_update}
\end{table*}

\subsubsection{Computation Cost Measured in Wall-Clock Time}
{
Test sample queue and multiple update epochs introduce additional computation overhead. To investigate the impact on efficiency, we measure the overall wall time as the time elapsed from the beginning to the end of test-time training, including all I/O overheads. The per-sample wall time is then calculated as the overall wall time divided by the number of test samples. 
We report the per-sample wall time (in seconds) for BN, TENT, SHOT, TTAC and TTAC++ in Tab.~\ref{tab:QueueTimeAnalysis} under different update epoch settings and without queue setting.
The Inference row indicates the per-sample wall-clock 
time in a single forward pass including data I/O overhead.
We observe that, under the same experiment setting, BN and TENT are more computational efficient, but TTAC++ is only 2 to 3 times more expensive than BN and TENT if no test sample queue is preserved (0.0090 v.s. 0.0030/0.0041) while the performance of TTAC++ w/o queue is still better than TENT (11.18 v.s. 13.48).
In summary, TTAC++
is able to strike a balance between computation efficiency and performance depending on how much computation resource is available. This suggests allocating a separate device for model weights update is only necessary when securing best performance is the priority.
}

\subsubsection{Evaluation of compatibility with Transformer Backbone}
In this section, we provide additional evaluation of TTAC++ with a transformer backbone, ViT~\cite{dosovitskiy2020vit}. In specific, we pre-train ViT on CIFAR10 clean dataset and then follow the sTTT protocol to do test-time training on CIFAR10-C testing set. The results are presented in Tab.~\ref{tab:ViT}. We report the average~(Avg) and standard deviation~(Std) of accuracy over all 15 categories of corruptions. Again, TTAC++ consistently outperform all competing methods with transformer backbone.

\subsubsection{Impact of Data Streaming Order}
The proposed sTTT protocols assumes test samples arrive in a stream and inference is made instantly on each test sample. The result for each test sample will not be affected by any following ones. In this section, we investigate how the data streaming order will affect the results. Specifically, we randomly shuffle all testing samples in CIFAR10-C for 10 times with different seeds and calculate the mean and standard deviation of test accuracy under sTTT protocol. The results in Tab.~\ref{tab:stream_order} suggest TTAC++ maintains consistent performance regardless of data streaming order.

% \begin{table*}[]
%     \centering
%     \caption{The performance of TTAC under different data streaming orders. }
%     \resizebox{\linewidth}{!}{
%         \begin{tabular}{l|cccccccccc|cc}
%         \toprule
%            Random Seed & 0 & 10 & 20 & 200 & 300 & 3000 & 4000 & 40000 & 50000 & 500000 & Avg \\
%         \midrule
%            Error ($\%$) & 10.01 & 10.06 & 10.05 & 10.29 & 10.20 & 10.03 & 10.31 & 10.36 & 10.37 & 10.13 & 10.18$\pm$0.13 \\
%         \bottomrule
%         \end{tabular}
%     }
%     \label{tab:stream_order}
% \end{table*}

\subsubsection{Sensitivity to Hyperparameters}
We evaluate the sensitivity to two thresholds during pseudo label filtering, namely the temporal smoothness threshold $\tau_{TC}$ and posterior threshold $\tau_{PP}$. In particular, $\tau_{TC}$ controls how much the maximal probability deviate from the historical exponential moving average~(ema). If the current value is lower than the ema below a threshold, we believe the prediction is not confident and the sample should be excluded from estimating target domain cluster. $\tau_{PP}$ controls the the minimal maximal probability and below this threshold is considered as not confident enough. We evaluate $\tau_{TC}$in the interval between 0 and -1.0 and $\tau_{PP}$ in the interval from 0.5 to 0.95 with results on CIFAR10-C level 5 glass blur corruption presented in Tab.~\ref{tab:Thresholds}. We draw the following conclusions on the evaluations. First, there is a wide range of hyperparameters that give stable performance, e.g. $\tau_{TC}\in[0.5,0.0.9]$ and $\tau_{PP}\in[-0.0001,-0.01]$. Second, when temporal consistency filtering is turn off, i.e. $\tau_{TC}=-1.0$, because the probability is normalized to between 0 and 1, the performance drops substantially, suggesting the necessity to apply temporal consistency filtering.

\begin{table}[!htb]
    \centering
    \caption{Evaluation of pseudo labeling thresholds on CIFAR10-C level 5 glass blur corruption. Numbers are reported as classification error (\%).}
        \resizebox{0.85\linewidth}{!}{
    \begin{tabular}{c|ccccccc}
    \toprule
        $\tau_{TC}\backslash \tau_{PP}$ & 0.5 & 0.6 & 0.7 & 0.8 & 0.9 & 0.95 \\
        \midrule
        0.0 & 23.03 & 22.26 & 21.96 & 22.50 & 21.14 & 28.55 \\
        -0.0001 & 20.03 & 20.53 & 20.45 & 20.40 & 19.49 & 27.00 \\
        -0.001 & 19.66 & 20.51 & 19.49 & 20.48 & \textbf{19.42} & 26.83 \\
        -0.01 & 20.71 & 20.78 & 20.73 & 20.65 & 20.29 & 27.58 \\
        -0.1 & 24.10 & 21.47 & 21.46 & 22.36 & 21.45 & 28.71 \\
        -1.0 & 30.75 & 24.08 & 23.40 & 24.33 & 22.21 & 28.77 \\
    \bottomrule
    \end{tabular}
    }
    \label{tab:Thresholds}
\end{table}

\subsubsection{Alternative Strategies for Updating Target Domain Clusters
}
In Sect.~\ref{sect:cluster_pl}, we presented target domain clustering through pseudo labeling. A temporal consistency approach is adopted to filter out confident samples to update target clusters. In this section, we discuss two alternative strategies for updating target domain clusters. Firstly, each target cluster can be updated with all samples assigned with respective pseudo label~(without Filtering). This strategy will introduce many noisy samples into cluster updating and potentially harm test-time feature learning. Secondly, we use a soft assignment of testing samples to each target cluster to update target clusters~(Soft Assignment). This strategy is equivalent to fitting a mixture of Gaussian through EM algorithm. We compare these two alternative strategies with our temporal consistency based filtering approach. The results are presented in Tab.~\ref{tab:target_clust_update}. We find the results with temporal consistency based filtering outperforms the other two strategies on 13 out of 15 categories of corruptions, suggesting pseudo label filtering is necessary for estimating more accurate target clusters.

\subsubsection{Alternative Design for Inferring Source Domain Distributions}
 In this work, we develop a solution to infer the source distributions in Sect.~\ref{sect:SFTTT}. Alternative to learning the distribution mean, an alternative solution is developed by re-scaling the classifier weight~\cite{ding2022source}. In this section, we compare the two options for estimating source domain statistics with results presented in Tab.~\ref{tab:estimated_source_distribution}.
 We conclude from the comparison that learning source domain statistics~(TTAC++) is clearly better than re-scaling classifier weights~\cite{ding2022source}. We attribute the advantage of TTAC++ to the fact that backbone features are obtained after ReLu activation, thus being all positive. The classifier weights are trained without any constraints and could have negative weights. The mismatch between classifier weights and backbone features might lead to inferior results by using re-scaled classifier weights as source domain distribution mean.
 % , we found that learned source distribution is more explainable and effective than directly re-scaling the classifier weight. One common situation, where the data features are all positive obtained after the ReLU~\cite{glorot2011deep} activation function while classifier weights may be negative, naturally contradicts the assumption that means and weights have the same direction. The specific experiments are conducted in Tab.~\ref{tab:estimated_source_distribution}.

\begin{table}[ht]
    \centering
    \caption{Comparing alternative methods to estimate source domain distributions.}%The choice of the mean of estimated source distribution. Weight denotes the re-scaling weight of the classifier, TTAC++(SF) leverages the learned prototypes, and TTAC++(SL) leverages the statistics calculated from source data. These results are averaged error rates above all level-5 corrupted sets on CIFAR10-C dataset. }
    \resizebox{0.45\linewidth}{!}{
    \begin{tabular}{c|c}
    \toprule
        Method &  Error (\%) \\
    \midrule
        Class. Weights~\cite{ding2022source} & 13.79\\
        TTAC++~(SF) & 11.62\\
    % \midrule
        % TTAC++(SL) & 9.78\\
    \bottomrule
    \end{tabular}
    }
    \label{tab:estimated_source_distribution}
\end{table}

\subsubsection{Limitations and Failure Cases}

Finally, we discuss the limitations of TTAC++ from two perspectives. First, we point out that TTAC++ requires backpropagation to update models at testing stage, therefore additional computation overhead is required. As shown in Tab.~\ref{tab:QueueTimeAnalysis}, TTAC++ is 2-5 times computationally more expensive than BN and TENT. However, contrary to usual expectation, BN and TENT are also very expensive compared with no adaptation at all. Eventually, most test-time training methods might require an additional device for test-time adaptation. 

We further discuss the limitations on test-time training under more severe corruptions. Specifically, we evaluate TENT, SHOT, TTAC, and TTAC++ under 1-5 levels of corruptions on CIFAR10-C with results reported in Tab.~\ref{tab:CorruptLevel}. We observe generally a drop of performance from 1-5 level of corruption. Despite consistently outperforming TENT and SHOT at all levels of corruptions, TTAC++'s performance at higher corruption levels are relatively worse, suggesting future attention must be paid to more severely corrupted scenarios. 

\begin{table}[ht]
    \caption{Classification error under different levels of snow corruption on CIFAR10-C dataset.}
    \centering
        \resizebox{0.7\linewidth}{!}{
    \begin{tabular}{c|ccccc}
    \toprule
        Level & 1 & 2 & 3 & 4 & 5 \\
    \midrule
        TEST & 9.46 & 18.34 & 16.89 & 19.31 & 21.93 \\
        TENT & 8.76 & 11.39 & 13.37 & 15.18 & 13.93 \\
        SHOT & 8.70 & 11.21 & 13.16 & 15.12 & 13.76 \\
        TTAC & 6.54 & 8.19 & 9.82 & 10.61 & 10.01 \\
        TTAC++ & \textbf{6.05} & \textbf{7.83} & \textbf{8.47} & \textbf{9.43} & \textbf{8.82}\\
    \midrule
    \end{tabular}
    }
    \label{tab:CorruptLevel}
\end{table}

\vspace{-0.5cm}

\section{Conclusion}
Test-time training~(TTT) tackles the realistic challenges of deploying domain adaptation on-the-fly. In this work, we are first motivated by the confused evaluation protocols for TTT and proposed two key criteria, namely modifying source training objective and sequential inference, to further categorize existing methods into four TTT protocols. Under the most realistic protocol, i.e. sequential test-time training (sTTT), we developed a test-time anchored clustering (TTAC) approach to align target domain features to the source ones. Unlike batchnorm and classifier prototype updates, anchored clustering allows all network parameters to be trainable, thus demonstrating stronger test-time training ability. We further proposed pseudo label filtering and an iterative update method to improve anchored clustering and save memory footprint respectively. When source domain distribution information is absent, we proposed to infer the distribution for anchored clustering through efficient gradient based optimization to achieve source-free sTTT. Finally, we incorporated self-training to update model weights with high confidence pseudo labels. We demonstrated self-training is particularly helpful with anchored clustering as regularization, and the improved model is referred to as TTAC++. Experiments on five datasets verified the effectiveness of TTAC++ under sTTT as well as other TTT protocols. We hope this work will serve as an in time taxonomy of TTT protocols and future works can be compared fairly under respective protocols.

\noindent\textbf{Acknowledgement:} This work was supported in part by the National Natural Science Foundation of China (NSFC) under Grant 62106078, Guangdong R\&D key project of China (No.:
2019B010155001), the Program for Guangdong Introducing Innovative and Enterpreneurial
Teams (No.: 2017ZT07X183), and A*STAR Career Development Award (Grant no. C210112059).
% \input{Supplementary.tex}

% \section{Appendix}

% \input{Appendix.tex}

% trigger a \newpage just before the given reference
% number - used to balance the columns on the last page
% adjust value as needed - may need to be readjusted if
% the document is modified later
%\IEEEtriggeratref{8}
% The "triggered" command can be changed if desired:
%\IEEEtriggercmd{\enlargethispage{-5in}}

% references section

% can use a bibliography generated by BibTeX as a .bbl file
% BibTeX documentation can be easily obtained at:
% http://mirror.ctan.org/biblio/bibtex/contrib/doc/
% The IEEEtran BibTeX style support page is at:
% http://www.michaelshell.org/tex/ieeetran/bibtex/
\bibliographystyle{IEEEtran}
% argument is your BibTeX string definitions and bibliography database(s)
\bibliography{IEEEabrv,./Reference}
%
% <OR> manually copy in the resultant .bbl file
% set second argument of \begin to the number of references
% (used to reserve space for the reference number labels box)
%\begin{thebibliography}{1}
%
%\bibitem{IEEEhowto:kopka}
%H.~Kopka and P.~W. Daly, \emph{A Guide to \LaTeX}, 3rd~ed.\hskip 1em plus
%  0.5em minus 0.4em\relax Harlow, England: Addison-Wesley, 1999.
%
%\end{thebibliography}

% biography section
% 
% If you have an EPS/PDF photo (graphicx package needed) extra braces are
% needed around the contents of the optional argument to biography to prevent
% the LaTeX parser from getting confused when it sees the complicated
% \includegraphics command within an optional argument. (You could create
% your own custom macro containing the \includegraphics command to make things
% simpler here.)
%\begin{IEEEbiography}[{\includegraphics[width=1in,height=1.25in,clip,keepaspectratio]{mshell}}]{Michael Shell}
% or if you just want to reserve a space for a photo:

\begin{IEEEbiography}
[{\includegraphics[width=1in, height=1.25in, clip, keepaspectratio]{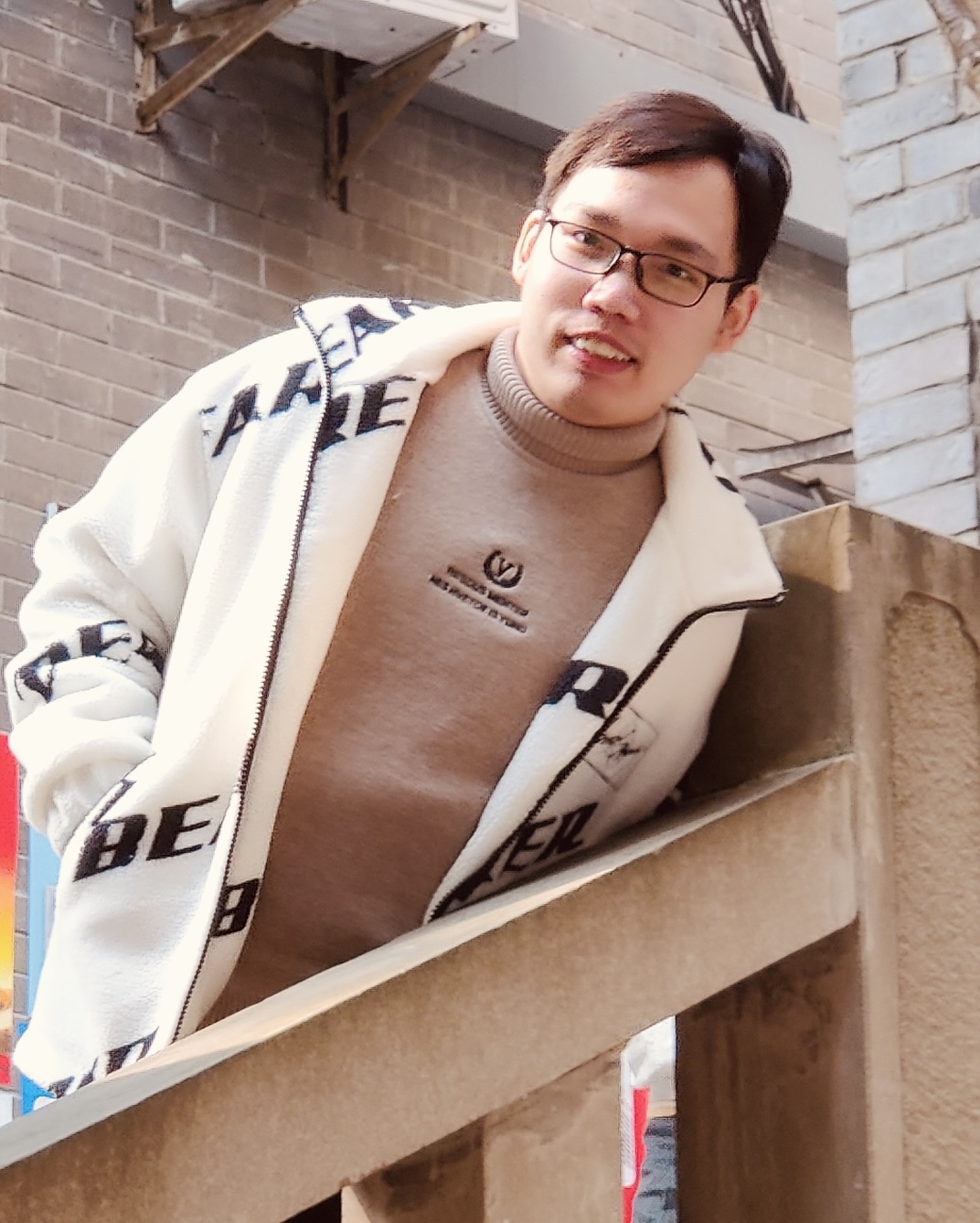}}]{Yongyi Su}
received the B.E. degree from South China University of Technology in 2021. He is currently working toward the PhD degree in the School of Electronic and Information Engineering, South China University of Technology, Guangzhou, China. His research interests mainly include 3D weakly supervised learning, domain adaptation, test-time training and robust learning.
\end{IEEEbiography}

\begin{IEEEbiography}[{\includegraphics[width=1in,height=1.25in,clip,keepaspectratio]{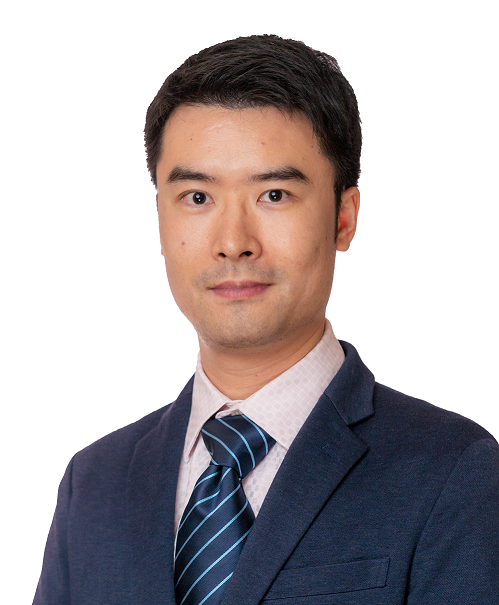}}]{Xun Xu}
received the B.E. degree from Sichuan
University, in 2010 and the PhD degree from Queen Mary University of London in 2016. He was a research fellow with National University of Singapore between 2016 and 2019. He is now with I2R, A*STAR.
His research interests include semi-supervised learning, domain adaptation, zero-shot learning with applications to 3D point cloud data.
\end{IEEEbiography}

\begin{IEEEbiography}[{\includegraphics[width=1in,height=1.25in,clip,keepaspectratio]{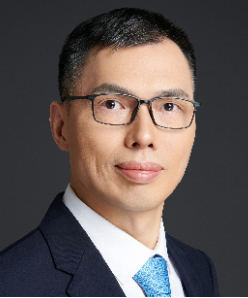}}]{Tianrui Li}
(SM'11) received the B.S., M.S., and Ph.D. degrees from Southwest Jiaotong University, Chengdu, China, in 1992, 1995, and 2002, respectively. He was a Post-Doctoral Researcher with Belgian Nuclear Research Centre, Mol, Belgium, from 2005 to 2006, and a Visiting Professor with Hasselt University, Hasselt, Belgium, in 2008; University of Technology, Sydney, Australia, in 2009; and University of Regina, Regina, Canada, in 2014. He is currently a Professor and the Director of the Key Laboratory of Cloud Computing and Intelligent Techniques, Southwest Jiaotong University. He has authored or co-authored over 300 research papers in refereed journals and conferences. His research interests include big data, machine learning, data mining, granular computing, and rough sets.
\end{IEEEbiography}

\begin{IEEEbiography}[{\includegraphics[width=1in,height=1.25in,clip,keepaspectratio]{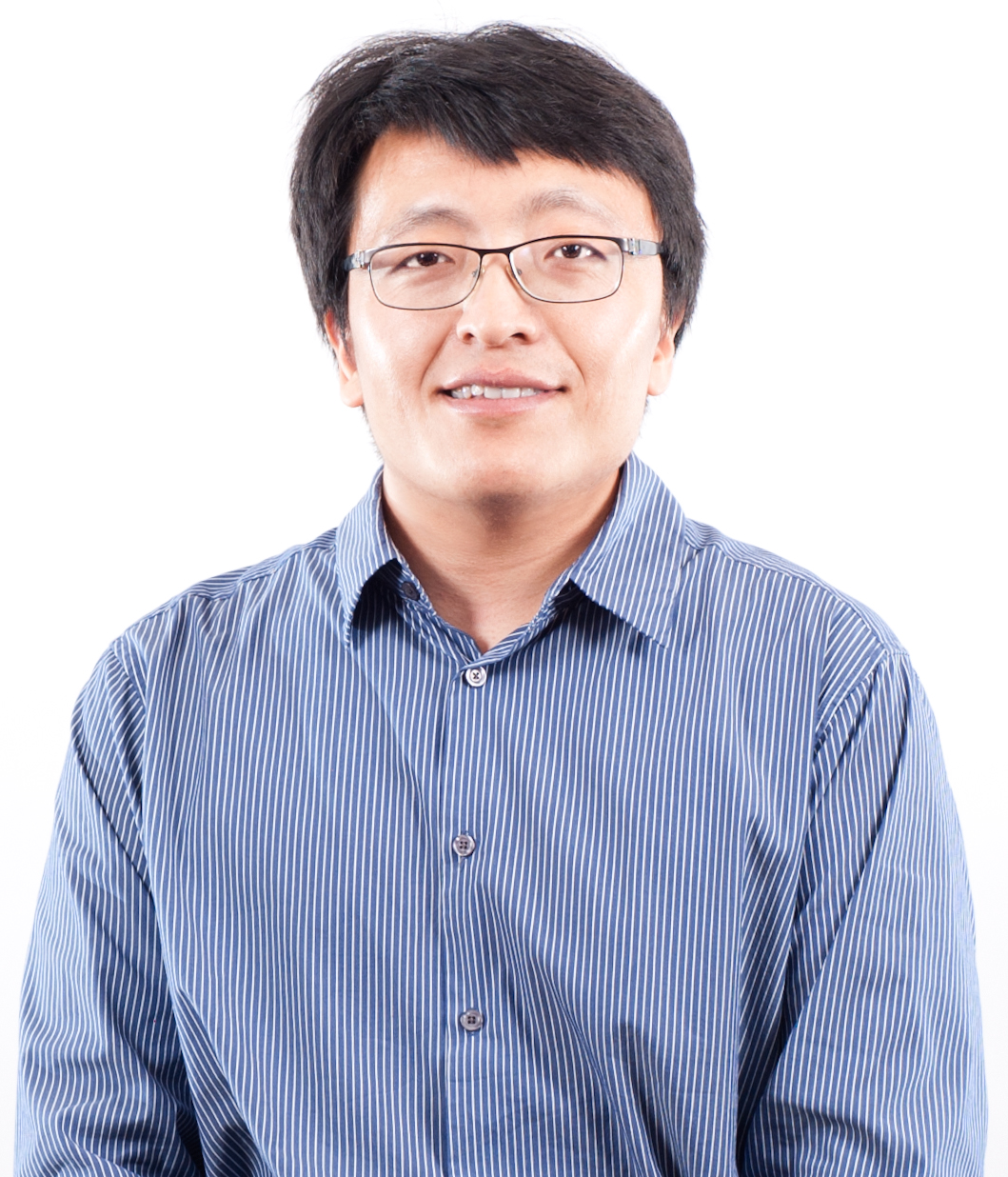}}]{Kui Jia}
 received the B.E. degree from Northwestern Polytechnic University, Xi’an, China, in
2001, the M.E. degree from the National University of Singapore, Singapore, in 2004, and
the Ph.D. degree in computer science from the
Queen Mary University of London, London, U.K.,
in 2007. He was with the Shenzhen Institute of
Advanced Technology of the Chinese Academy
of Sciences, Shenzhen, China, Chinese University of Hong Kong, Hong Kong, the Institute
of Advanced Studies, University of Illinois at
Urbana-Champaign, Champaign, IL, USA, and the University of Macau,
Macau, China. He is currently a Professor with the School of Electronic
and Information Engineering, South China University of Technology,
Guangzhou, China. His recent research focuses on theoretical deep
learning and its applications in vision and robotic problems, including
deep learning of 3D data and deep transfer learning.
\end{IEEEbiography}

%\begin{IEEEbiography}[{\includegraphics[width=1in,height=1.25in,clip,keepaspectratio]{./Figure/Biography/Zhu}}]{Ce Zhu}
%
%\end{IEEEbiography}

% You can push biographies down or up by placing
% a \vfill before or after them. The appropriate
% use of \vfill depends on what kind of text is
% on the last page and whether or not the columns
% are being equalized.

%\vfill

% Can be used to pull up biographies so that the bottom of the last one
% is flush with the other column.
%\enlargethispage{-5in}

% that's all folks
\end{document}